\documentclass{article}
\usepackage{style/unites}
\usepackage{XCharter}
\usepackage[scaled=1.1]{zlmtt} 

\usepackage[utf8]{inputenc}
\usepackage[T1]{fontenc}
\usepackage{microtype}

\usepackage{amsmath}
\usepackage{amssymb}
\usepackage{amsfonts}
\usepackage{amsthm}
\usepackage{mathtools}
\usepackage{mathrsfs}
\usepackage{physics}
\usepackage{braket}
\usepackage{slashed}
\usepackage{nicefrac}
\usepackage{textcomp}
\usepackage{dsfont}
\usepackage{bbm}
\usepackage{bm}

\usepackage{graphicx}
\usepackage{subcaption}
\usepackage[export]{adjustbox}
\usepackage{float}
\usepackage{booktabs}
\usepackage{dcolumn}
\newcolumntype{d}[1]{D{.}{.}{#1}}
\usepackage{bigstrut, tabularx, multirow, makecell, diagbox}
\usepackage{colortbl}
\usepackage{tabularray}
\UseTblrLibrary{booktabs}
\usepackage{threeparttable}
\usepackage{tablefootnote}
\usepackage{fontawesome5}

\usepackage{placeins}
\usepackage{caption}
\usepackage{footnote}
\usepackage{enumitem}
\usepackage{multicol}
\usepackage{xspace}
\usepackage{titletoc}
\usepackage{titlesec}
\usepackage[bottom]{footmisc}
\usepackage{setspace}

\usepackage{wrapfig}
\usepackage{tikz}
\usepackage{quantikz}
\usepackage{dashbox}
\usepackage{mdframed}
\usepackage{marvosym}
\usepackage{pifont}
\usepackage{CJK}
\usepackage{url}

\usepackage[table,x11names]{xcolor}
\usepackage[most]{tcolorbox}
\tcbuselibrary{breakable}
\usetikzlibrary{decorations.pathreplacing, fit}

\definecolor{primaryblue}{HTML}{0066CC}
\definecolor{accentcyan}{HTML}{00D4AA}
\definecolor{warmorange}{HTML}{FF6B35}
\definecolor{deepgray}{HTML}{2C3E50}
\definecolor{lightgray}{HTML}{F8F9FA}
\definecolor{gradientstart}{HTML}{667eea}
\definecolor{gradientend}{HTML}{764ba2}

\definecolor{citecolor}{HTML}{0071bc}
\definecolor{citeblue}{RGB}{0, 113, 188}
\definecolor{linkcolor}{HTML}{9A4D92}
\definecolor{firebrick}{rgb}{0.698,0.133,0.133}

\definecolor{paleviolet}{HTML}{E1EEFC}
\definecolor{CarolinaUltraLight}{HTML}{E7F4FC}
\definecolor{lightgrey}{RGB}{247, 247, 247}
\definecolor{shadecolor}{HTML}{EFEFEF}
\definecolor{lightyellow}{rgb}{1.0, 0.95, 0.7}
\definecolor{lightblue}{rgb}{0.90, 0.95, 1.0}
\definecolor{light-gray}{gray}{0.95}

\definecolor{darkgrey}{rgb}{0.5, 0.5, 0.5}
\definecolor{darkgreen}{rgb}{0, 0.5, 0}
\definecolor{mydarkblue}{rgb}{0,0.08,0.45}
\definecolor{mydarkblue2}{rgb}{0.133, 0.133, 0.698}
\definecolor{echodrk}{HTML}{0099cc}
\definecolor{mymauve}{rgb}{0.58,0,0.82}
\definecolor{midnightblue}{rgb}{0.1,0.1,0.44}
\definecolor{oxfordblue}{rgb}{0.0,0.13,0.28}
\definecolor{prussianblue}{rgb}{0.0,0.19,0.33}
\definecolor{coolteal}{rgb}{0, 0.45, 0.45}
\definecolor{olive}{rgb}{0.1, 0.3, 0}
\definecolor{mypurple}{rgb}{0.5,0,0.5}
\definecolor{almond}{rgb}{0.94, 0.87, 0.8}

\definecolor{blue_ampEncoding}{HTML}{DAE8FC}
\definecolor{green_encoder}{HTML}{D5E8D4}
\definecolor{purple_decoder}{HTML}{E1D5E7}
\definecolor{yellow_measure}{HTML}{FFF2CC}
\definecolor{gray_block}{HTML}{F5F5F5}
\definecolor{pink_dru}{HTML}{FAD9D5}
\definecolor{orange_v}{HTML}{FAD7AC}

\definecolor{colorA}{rgb}{1,0,0}
\definecolor{colorB}{rgb}{0,0.3,1}
\definecolor{colorC}{rgb}{0.9,0.8,0.2}
\definecolor{colorD}{rgb}{0,0.65,0}
\definecolor{lesslightgray}{rgb}{0.5,0.5,0.5}
\definecolor{fundamental}{RGB}{55, 110, 111}
\definecolor{Gred}{RGB}{219, 50, 54}
\definecolor{ToCgreen}{RGB}{0, 128, 0}
\definecolor{Sepia}{RGB}{112, 66, 20}
\definecolor{Dblue}{rgb}{0,0.08,0.45}
\definecolor{Blue}{rgb}{0, 0, 0.8}
\definecolor{blue}{rgb}{0,0,1}
\definecolor{UNCblue!10}{rgb}{0.84,0.91,0.98}
\definecolor{RowAlt}{rgb}{0.98,0.98,0.99}

\definecolor{CarolinaBlue}{HTML}{7BAFD4}        
\definecolor{CarolinaLightBlue}{HTML}{B3D4E5}   
\definecolor{CarolinaUltraLight}{HTML}{E8F4F8}  
\definecolor{CarolinaText}{HTML}{1C2B33}        

\usepackage[pagebackref=true,breaklinks=true,colorlinks,hyperfootnotes=false]{hyperref}
\hypersetup{
  colorlinks,
  citecolor=citeblue,
  linkcolor=firebrick,
  urlcolor=firebrick
}
\usepackage[nameinlink,capitalize,noabbrev]{cleveref}

\titlespacing\section{0pt}{4pt plus 4pt minus 2pt}{-2pt plus 2pt minus 2pt}
\titlespacing\subsection{0pt}{2pt plus 4pt minus 2pt}{-2pt plus 2pt minus 2pt}
\titlespacing\subsubsection{0pt}{2pt plus 4pt minus 2pt}{-2pt plus 2pt minus 2pt}

\makeatletter
\def\th@remark{%
  \thm@headfont{\bfseries}%
  \normalfont 
  \thm@preskip\topsep \divide\thm@preskip\tw@
  \thm@postskip\thm@preskip
}
\makeatother

\theoremstyle{definition}

\tcolorboxenvironment{theorem}{
  breakable,
  colback=black!10,
  colframe=white,
  width=\linewidth, 
  enlarge left by=0pt,
  enlarge right by=0pt,
  boxsep=5pt,
  boxrule=0pt,
  left=0pt,right=0pt,top=0pt,bottom=0pt,
  arc=8pt,
  before skip=\topsep,
  after skip=\topsep
}

\tcolorboxenvironment{lemma}{
  breakable,
  colback=black!10,
  colframe=white,
  width=\linewidth,
  enlarge left by=0pt,
  enlarge right by=0pt,
  boxsep=5pt,
  boxrule=0pt,
  left=0pt,right=0pt,top=0pt,bottom=0pt,
  arc=8pt,
  before skip=\topsep,
  after skip=\topsep
}

\tcolorboxenvironment{corollary}{
  breakable,
  colback=black!10,
  colframe=white,
  width=\linewidth,
  enlarge left by=0pt,
  enlarge right by=0pt,
  boxsep=5pt,
  boxrule=0pt,
  left=0pt,right=0pt,top=0pt,bottom=0pt,
  arc=8pt,
  before skip=\topsep,
  after skip=\topsep
}

\tcolorboxenvironment{proposition}{
  breakable,
  colback=black!10,
  colframe=white,
  width=\linewidth,
  enlarge left by=0pt,
  enlarge right by=0pt,
  boxsep=5pt,
  boxrule=0pt,
  left=0pt,right=0pt,top=0pt,bottom=0pt,
  arc=8pt,
  before skip=\topsep,
  after skip=\topsep
}

\tcolorboxenvironment{definition}{
  breakable,
  colback=black!10,
  colframe=white,
  width=\linewidth,
  enlarge left by=0pt,
  enlarge right by=0pt,
  boxsep=5pt,
  boxrule=0pt,
  left=0pt,right=0pt,top=0pt,bottom=0pt,
  arc=8pt,
  before skip=\topsep,
  after skip=\topsep
}

\tcolorboxenvironment{assumption}{
  breakable,
  colback=black!10,
  colframe=white,
  width=\linewidth,
  enlarge left by=0pt,
  enlarge right by=0pt,
  boxsep=5pt,
  boxrule=0pt,
  left=0pt,right=0pt,top=0pt,bottom=0pt,
  arc=8pt,
  before skip=\topsep,
  after skip=\topsep
}

\tcolorboxenvironment{claim}{
  breakable,
  colback=black!10,
  colframe=white,
  width=\linewidth,
  enlarge left by=0pt,
  enlarge right by=0pt,
  boxsep=5pt,
  boxrule=0pt,
  left=0pt,right=0pt,top=0pt,bottom=0pt,
  arc=8pt,
  before skip=\topsep,
  after skip=\topsep
}

\tcolorboxenvironment{problem}{
  breakable,
  colback=black!10,
  colframe=white,
  width=\linewidth,
  enlarge left by=0pt,
  enlarge right by=0pt,
  boxsep=5pt,
  boxrule=0pt,
  left=0pt,right=0pt,top=0pt,bottom=0pt,
  arc=8pt,
  before skip=\topsep,
  after skip=\topsep
}

\tcolorboxenvironment{question}{
  breakable,
  colback=black!10,
  colframe=white,
  width=\linewidth,
  enlarge left by=0pt,
  enlarge right by=0pt,
  boxsep=5pt,
  boxrule=0pt,
  left=0pt,right=0pt,top=0pt,bottom=0pt,
  arc=8pt,
  before skip=\topsep,
  after skip=\topsep
}

\newtcolorbox{titleblock}{
  enhanced,
  frame hidden,
  colback=CarolinaUltraLight,
  colframe=CarolinaUltraLight,
  boxrule=0pt,
  arc=10pt,
  left=14pt,
  right=14pt,
  top=14pt,
  bottom=14pt,
  width=\linewidth,
  before skip=12pt plus 4pt,
  after skip=12pt plus 4pt,
  grow to left by=1.5pt,
  grow to right by=1.5pt,
  before upper={
    \setlength{\parindent}{0cm}
    \setlength{\parskip}{0.5cm}
  }
}

\crefname{theorem}{Theorem}{Theorems}
\crefname{proposition}{Proposition}{Propositions}
\crefname{lemma}{Lemma}{Lemmas}
\crefname{corollary}{Corollary}{Corollaries}
\crefname{definition}{Definition}{Definitions}
\crefname{assumption}{Assumption}{Assumptions}
\crefname{remark}{Remark}{Remarks}
\crefname{problem}{Problem}{Problems}
\crefname{property}{Property}{property}
\crefname{question}{Question}{Questions}

\numberwithin{equation}{section}
\numberwithin{theorem}{section}
\numberwithin{proposition}{section}
\numberwithin{definition}{section}
\numberwithin{lemma}{section}
\numberwithin{assumption}{section}
\numberwithin{remark}{section}

\def\1{\bm{1}}

\makeatletter
\let\save@mathaccent\mathaccent
\newcommand*\if@single[3]{%
    \setbox0\hbox{${\mathaccent"0362{#1}}^H$}%
    \setbox2\hbox{${\mathaccent"0362{\kern0pt#1}}^H$}%
    \ifdim\ht0=\ht2 #3\else #2\fi
}
\newcommand*\rel@kern[1]{\kern#1\dimexpr\macc@kerna}
\newcommand*\widebar[1]{\@ifnextchar^{{\wide@bar{#1}{0}}}{\wide@bar{#1}{1}}}
\newcommand*\wide@bar[2]{\if@single{#1}{\wide@bar@{#1}{#2}{1}}{\wide@bar@{#1}{#2}{2}}}
\newcommand*\wide@bar@[3]{%
    \begingroup
    \def\mathaccent##1##2{%
        \let\mathaccent\save@mathaccent
        \if#32 \let\macc@nucleus\first@char \fi
        \setbox\z@\hbox{$\macc@style{\macc@nucleus}_{}$}%
        \setbox\tw@\hbox{$\macc@style{\macc@nucleus}{}_{}$}%
        \dimen@\wd\tw@
        \advance\dimen@-\wd\z@
        \divide\dimen@ 3
        \@tempdima\wd\tw@
        \advance\@tempdima-\scriptspace
        \divide\@tempdima 10
        \advance\dimen@-\@tempdima
        \ifdim\dimen@>\z@ \dimen@0pt\fi
        \rel@kern{0.6}\kern-\dimen@
        \if#31
        \overline{\rel@kern{-0.6}\kern\dimen@\macc@nucleus\rel@kern{0.4}\kern\dimen@}%
        \advance\dimen@0.4\dimexpr\macc@kerna
        \let\final@kern#2%
        \ifdim\dimen@<\z@ \let\final@kern1\fi
        \if\final@kern1 \kern-\dimen@\fi
        \else
        \overline{\rel@kern{-0.6}\kern\dimen@#1}%
        \fi
    }%
    \macc@depth\@ne
    \let\math@bgroup\@empty \let\math@egroup\macc@set@skewchar
    \mathsurround\z@ \frozen@everymath{\mathgroup\macc@group\relax}%
    \macc@set@skewchar\relax
    \let\mathaccentV\macc@nested@a
    \if#31
    \macc@nested@a\relax111{#1}%
    \else
    \def\gobble@till@marker##1\endmarker{}%
    \futurelet\first@char\gobble@till@marker#1\endmarker
    \ifcat\noexpand\first@char A\else
    \def\first@char{}%
    \fi
    \macc@nested@a\relax111{\first@char}%
    \fi
    \endgroup
    }
\makeatother
\let\bar\widebar

\DeclareMathAlphabet{\mathsfit}{\encodingdefault}{\sfdefault}{m}{sl}
\SetMathAlphabet{\mathsfit}{bold}{\encodingdefault}{\sfdefault}{bx}{n}

\let\tilde\widetilde
\let\hat\widehat

\renewcommand{\arraystretch}{1.15}
\usepackage{hyperref}
\usepackage[dvipsnames]{xcolor}  
\hypersetup{
    colorlinks=true, 
    linkcolor=Blue,
    filecolor=Blue,      
    urlcolor=Blue,
    citecolor=Blue 
    }
\usepackage{url}
\usepackage[utf8]{inputenc}
\usepackage[T1]{fontenc}
\usepackage{booktabs}
\usepackage{amsfonts,amsmath,amssymb}
\usepackage{nicefrac}
\usepackage{microtype}
\usepackage{graphicx}
\usepackage{subcaption}
\usepackage{xcolor}
\usepackage{wrapfig}
\usepackage{multirow}
\usepackage[table]{xcolor}

\newcommand{\method}{\texttt{MOSAIC}}

\usepackage{xcolor}

\begin{document}

\makeatletter
\def\blfootnote{\gdef\@thefnmark{}\@footnotetext}
\makeatother

\makeatletter
\pagestyle{fancy}
\fancyhf{}
\renewcommand{\headrulewidth}{1pt}
\chead{\small\bf Cross-Modal Latent Influence: Patch-Level Training Sample Attribution for Vision-Language Models
}
\cfoot{\thepage}
\thispagestyle{fancy}
\makeatother

\makeatletter
\def\icmldate#1{\gdef\@icmldate{#1}}
\icmldate{\today}
\makeatother

\makeatletter
\fancypagestyle{fancytitlepage}{
  \fancyhead{}
  \lhead{\includegraphics[height=0.8cm]{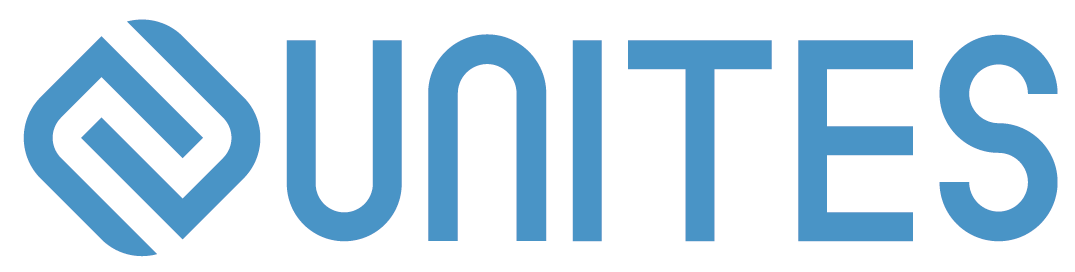}}
  \rhead{\it \@icmldate}
  \cfoot{}
}
\makeatother

\thispagestyle{fancytitlepage}

\vspace*{0.5em}

\noindent
\begin{titleblock}
    {\setlength{\parskip}{0cm}
     \raggedright
     {\setstretch{1.2}
      \LARGE\bfseries
      Cross-Modal Latent Influence: Patch-Level Training Sample Attribution for Vision-Language Models
      \par}
    }
    \vskip 0.2cm
    
    \begin{icmlauthorlist}
\mbox{Laura Yao$^{\,1\,\dagger\,\textrm{\Letter}\,}$},
\mbox{Shixing Yu$^{\,2}\,$}, 
\mbox{Jie Peng$^{\,3\,}$}, 
\mbox{Promit Ghosal$^{\,4\,}$}, 
\mbox{Kyra Gan$^{\,5\,*\,}$}
and \mbox{Tianlong Chen$^{\,1\,*\,}$}
\end{icmlauthorlist}

$^{1\,}$UNITES Lab, University of North Carolina at Chapel Hill  \quad $^{2\,}$Electrical and Computer Engineering, Cornell Tech \quad $^{3\,}$Foci Labs \quad $^{4\,}$Department of Statistics, University of Chicago \quad $^{5\,}$Operations Research and Industrial Engineering, Cornell Tech

$^{\dagger}$ Project Lead \quad
$^{*}$ Co-corresponding Author
    \vskip 0.2cm
    
    Explaining a vision--language model (VLM) prediction requires tracing training-example influence to the multimodal evidence that mediates it, yet existing attribution methods largely operate only at the sample level.
Recent latent-mediated influence methods provide feature-level attribution in language models, but lack the multimodal and spatial structure needed to ground influence in VLMs.
We introduce \method{} (Multimodal Orthogonalized Sparse Autoencoders for Influence Characterization).
It learns a shared sparse image--text latent space whose concepts are 1) sparse within each modality, 2) supported on contiguous image regions, 3) aligned to the phrases that describe them, and 4) approximately decorrelated. 
It then applies EK-FAC to decompose training-example influence through these concepts and localize their contributions to specific text phrases and image regions.
We evaluate \method{} on Qwen and Gemma across captioning, chest X-ray understanding, and visual question answering. 
The learned features are
more localized and coherent without degrading task behavior, and ablating influence-ranked latents changes model outputs more than ablating latents ranked by activation or frequency.
We find that visual evidence can substantially reorganize training-data attribution while preserving more of the shared latent structure. \method{} thus connects \emph{where model behavior comes from} with \emph{what multimodal evidence carries that influence}.

\end{titleblock}

\blfootnote{%
$^{\textrm{\Letter}}$ Contact: lyao@cs.unc.edu
\\[2.5em]
\ifcsname @icmlpreprint\endcsname
  \textit{\csname @icmlpreprint\endcsname}%
\fi
}

\section{Introduction}

\begin{wrapfigure}{r}{0.5\textwidth}
    \vspace{-23pt}
    \centering
    \includegraphics[width=\linewidth]{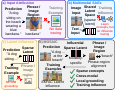}
    \vspace{-23pt}
    \caption{(a) Input attribution localizes evidence but lacks training-data specific attribution.
    (b) Multimodal SAEs expose internal concepts without tracing their training origins.
    (c) Latent influence traces training examples through features but lacks multimodal grounding.
    (d) \method{} connects influential training examples to sparse multimodal concepts and localized evidence.}
    \label{fig:intro}
    \vspace{-17pt}
\end{wrapfigure}

Understanding a visual prediction requires more than identifying salient pixels or patches:
it requires linking the visual evidence a model relies on to the training examples that shaped that reliance.
Vision attribution methods can localize prediction-relevant regions \citep{selvaraju2017gradcam,chefer2021generic} (Fig.~\ref{fig:intro}a), while data-attribution methods trace predictions to influential training examples \citep{koh2017influence,park2023trak,grosse2023studying}. 
Yet, these two views are rarely connected, especially for
vision--language models (VLMs).
Even when data attribution successfully identifies an influential training example, it treats the image as a monolithic unit, answering only \emph{which} example mattered, not \emph{how} it mattered. To truly audit a VLM, we must drill down from the entire training sample to the specific concepts within it that actually drove the prediction.

%
VLMs make this finer-grained, concept-level attribution uniquely tractable. Unlike language-only models, influential features can be inspected along both \emph{semantic} and \emph{spatial} dimensions: a feature associated with a phrase can be directly checked against the image region on which it activates. This dual-sided inspection reveals not just the influential example, but \emph{what concept carries its influence and where that concept appears}. This yields the multimodal attribution chain:
$
\text{training example}
\leftrightarrow
\text{multimodal concept}
\leftrightarrow
\text{phrase / image region}.
$

Recent work on latent-mediated influence provides an important step toward such explanations in language models \citep{yu2026correcting}. It decomposes training-example influence through sparse latent features and maps those features back to tokens (Fig.~\ref{fig:intro}c). 
Extending this formulation to VLMs, however, is not straightforward.
Since
VLM representations jointly encode image and text tokens, visual evidence carries explicit spatial structure rather than forming a linear sequence; further,
the two modalities can have highly imbalanced token counts while still sharing latent directions.
Existing multimodal SAEs, in contrast, can provide interpretable image--text features \citep{pach2025sparse,shen2025vlsae,shu2026joint,liao2026structured}, but are not tailored for training-data attribution and auditing (Fig.~\ref{fig:intro}b).

We introduce \method{} (Multimodal Orthogonalized Sparse Autoencoders for Influence Characterization), a framework for grounded latent-mediated influence in VLMs (Fig.~\ref{fig:intro}d)
by learning a latent representation whose features are both spatially grounded and cross-modally aligned, then attributing training-example influence through those features.
\method{} first learns a shared TopK dictionary over image and text representations while preserving their distinct structure. 
To ensure that visual features correspond to coherent regions rather than scattered patches, it groups patches into regions, then imposes group and exclusive sparsity. To ensure that features are comparable across modalities, it aligns pooled phrase and region responses in the shared latent space and decorrelates latent activity, yielding a better-separated basis for attribution. 
Second, given this representation, an EK-FAC-based~\citep{grosse2023studying}
influence estimator decomposes each training example's effect over individual sparse features and projects those contributions back to the corresponding text phrases and image regions:
$
\text{prediction}
\leftrightarrow
\text{training example}
\leftrightarrow
\text{sparse latent}
\leftrightarrow
\text{phrase / image region}.
$

\method{} (as illustrated in Fig.~\ref{fig:method_overview}) enables questions that language-only attribution cannot pose. Do influential features correspond to localized cross-modal evidence? Are they causally important to the prediction? And how does visual evidence change the training examples through which a prediction is supported?  
We study these questions across two VLM architectures (Qwen and Gemma) and three tasks: image captioning, chest X-ray understanding, and visual question answering.

Our experiments show that
1) the structured latents are spatially localized and semantically grounded,
2) influence ranking identifies behaviorally consequential features,
and 3) visual evidence substantially reshapes training-data attribution.
First, the structured latent space yields substantially more localized multimodal features while largely preserving task performance: on GQA with Qwen, visual footprint (the fraction of image patches on which a feature activates) drops from $0.688$ to $0.120$, indicating that visual features concentrate on substantially smaller regions, with similar gains across tasks/backbones.

Second, influence-ranked latents are markedly more behaviorally consequential than activation-based alternatives: on GQA with Gemma, removing the top 256 influence-ranked latents flips $38.3\%$ of predictions versus $8.3\%$ under activation ranking, with the same trend across datasets and models. Together, these results support the learned latent decomposition as a meaningful basis for attribution.

Finally, visual evidence substantially reshapes training-data attribution. Removing the image drives training-example rank correlation as low as $\rho=-0.797$, while latent rankings remain positively correlated at up to $\rho=0.795$. Thus, the influential examples can change dramatically even as substantially more of the underlying latent feature structure is preserved.


In summary, our contributions are threefold: (1) a \textbf{grounded training-data attribution framework for VLMs} that links training examples to localized phrases and image regions through a shared sparse latent basis;
(2) an \textbf{attribution-oriented multimodal sparse autoencoder (SAE)} whose features are spatially coherent and cross-modally aligned; 
and (3) an \textbf{empirical characterization} showing that visual evidence can substantially reorganize training-data attribution while latent structure remains more stable and that the resulting features are behaviorally consequential and semantically grounded.
\section{Related Work}

\textbf{Training-data attribution and latent influence.}
Training-data attribution methods, from influence functions to scalable approaches such as TRAK and approximations like EK-FAC
\citep{koh2017influence,park2023trak,grosse2023studying}, estimate how training examples affect model predictions.
Recent work extends attribution toward more scalable or concept-level analyses, including sparse-recovery-based attribution, semantic concept influence, and concept-level attribution in generative models
\citep{dagli2026stride,kowal2026conceptinfluence,park2026concepttrak}.
Most closely related, \citet{yu2026correcting} decomposes example influence through approximately orthogonal SAE features, exposing the latent features that mediate a prediction.
We extend this view to VLMs, where influence must be resolved jointly over linguistic and spatially organized visual representations, enabling a hierarchy from influential training examples to multimodal concepts and localized evidence.

\textbf{Visual and multimodal attribution.}
Visual attribution methods localize prediction-relevant evidence using gradients, attention, or relevance propagation
\citep{selvaraju2017gradcam,abnar2020quantifying,chefer2021generic}.
Recent work further studies latent visual attribution, modality reliance, grounding, causal intervention, and mechanistic tracing in VLMs
\citep{dreyer2025attributing,parcalabescu2025vision,geigle-etal-2024-object,li2025visual,
wang2025vseam,park2026save}.
Together, these approaches identify or manipulate prediction-relevant visual evidence, modality contributions, or internal mechanisms, but do not connect them to the training examples that shaped the prediction.
Since attention itself need not provide a faithful explanation
\citep{jain2019attention,wiegreffe2019attention}, we use frozen VLM attention only to construct phrase--region correspondences; attribution is instead determined by training-data influence through the learned latent representation.

\textbf{Sparse multimodal representations.}
SAEs decompose dense activations into sparse, interpretable features
\citep{huben2024sparse,gao2024scaling}.
Recent multimodal extensions study shared image--text features, cross-modal alignment, hierarchical and compositional visual concepts, and spatially structured representations
\citep{pach2025sparse,shen2025vlsae,shu2026joint,nasirisarvi2026sparc,
lou2025saev,zaigrajew2025clip,zhao2026cascaded,lanza2026multimodal,liao2026structured}.
However, these representations are not designed to mediate example-level training-data influence.
In contrast, we learn a shared, position-resolved image--text latent basis with spatial coherence and phrase--region alignment, and decompose training-example influence through this basis to connect predictions, training examples, multimodal concepts, and localized visual and linguistic evidence.

\vspace{-2.3mm}
\section{Method}

\label{sec:method}
This section describes \method{} (Fig.~\ref{fig:method_overview}). We first review the latent-mediated influence formulation on which our method builds and define the shared SAE representation (\S\ref{sec:background}). We then introduce spatial constraints on visual features (\S\ref{sec:spatial}) and phrase region alignment (\S\ref{sec:alignment}). We aggregate latent attribution scores over image regions and text phrases (\S\ref{sec:multimodal_influence}), using EK-FAC for scalable inverse curvature computation (\S\ref{sec:ekfac}). The SAE is optimized in two stages: learning a reconstructive dictionary, followed by spatial and cross-modal shaping. Full SAE configuration, layer placement, and optimization details are provided in Appendix~\ref{app:implementation}.


\subsection{Background and Multimodal Latent Basis}
\label{sec:background}

Following \citet{yu2026correcting}, we split the VLM at layer $l$ as
$h_{\theta}=h_{\theta_2}\circ h_{\theta_1}$ and insert an SAE there.
For example $i$, let
$R_i\in\mathbb{R}^{T_i\times h}$ denote its sparse SAE activations, where
$r_{itj}$ is the activation of latent $j$ at sequence position $t$.
The SAE decoder reconstructs the intermediate representation from $R_i$, which
is then passed to the upper model $h_{\theta_2}$.

\begin{wrapfigure}{r}{0.6\textwidth}
    \vspace{-10mm}
    \centering
    \includegraphics[width=\linewidth]{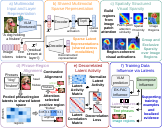}
   \vspace{-25pt}
    \caption{Overview of \method{}. (a) Intermediate VLM image and text representations. (b) Shared TopK SAE with a common sparse dictionary. (c) Spatially structured sparsity for region-coherent visual features. (d) Contrastive phrase--region alignment using frozen VLM attention. (e) Latent decorrelation for a better-separated attribution space. (f) EK-FAC influence tracing predictions to training examples, latents, phrases, and image regions.}
    \label{fig:method_overview}
    \vspace{-10mm}
\end{wrapfigure}

For training example $z_i$ and target example $z_\star$, define their gradients
through the SAE-mediated forward path as
$g_i(R_i)
=
\nabla_{\theta_2}\ell(z_i;\theta_2,R_i),
\quad
g_\star
=
\nabla_{\theta_2}\ell(z_\star;\theta_2,R_\star).$
Given curvature approximation $C_{\theta_2}$, the estimated effect of
upweighting $z_i$ on the target loss is
$I(z_i\!\rightarrow\!z_\star)
=
-g_\star^\top C_{\theta_2}^{-1}g_i(R_i).$

To decompose this example-level influence over latent features, define
$P_i(R_i)
=
g_\star^\top C_{\theta_2}^{-1}g_i(R_i).$
The contribution of latent $j$ at position $t$ is
$A_{itj}
=
-r_{itj}\frac{\partial P_i(R_i)}{\partial r_{itj}},
\quad
A_{ij}
=
\sum_{t=1}^{T_i}A_{itj}.$
Here, the target gradient and curvature estimate are fixed while the training
example's latent representation is varied, identifying which active features
carry its influence on $z_\star$. This
activation-weighted derivative gives a first-order attribution relative to an
inactive-feature baseline, so inactive latents receive zero attribution.
To make this decomposition multimodal and spatially localizable, we construct a latent basis over visual and linguistic representations. As illustrated
in Fig.~\ref{fig:method_overview}a--b, at layer $l$ the VLM represents example
$i$ as a sequence of image- and text-token hidden states,
$X_i^{l}
=
[x^{l}_{i1}\cdots x^{l}_{iT_i}]^\top
\in\mathbb{R}^{T_i\times d},$
where $x_{it}^l\in\mathbb{R}^d$ is the hidden representation of one image or
text token, $T_i$ is the sequence length, and $d$ is the VLM hidden dimension.
We denote the image and text positions with
$\mathcal{P}_{\mathrm{img}}$ and $\mathcal{P}_{\mathrm{txt}}$.

We apply the same TopK SAE position-wise to both modalities:
$a_{it}
=
W_{\mathrm{enc}}(x_{it}^l-b_{\mathrm{pre}})+b_{\mathrm{enc}},
\quad
r_{it}
=
\operatorname{ReLU}(\operatorname{TopK}_k(a_{it})),
\quad
\hat{x}_{it}^l
=
W_{\mathrm{dec}}r_{it}+b_{\mathrm{pre}}.$
Here, $a_{it},r_{it}\in\mathbb{R}^h$ are the latent pre-activation and sparse
code, respectively, and $\hat{x}_{it}^l\in\mathbb{R}^d$ reconstructs the
original hidden state. $\operatorname{TopK}_k$ retains the $k$ largest entries
before ReLU, yielding at most $k$ active features per position. Stacking
$r_{it}$ across positions gives $R_i$ above.

The shared encoder and decoder give image and text positions a common latent dictionary, while visual activations are further shaped by the spatial constraints in \S\ref{sec:spatial}. Finally, because redundant latent coordinates make feature-level influence
difficult to interpret, we encourage approximately decorrelated latent activity.
This regularizes latent activation patterns rather than enforcing orthogonality
of the overcomplete decoder dictionary. The exact decorrelation objective and
SAE training details are provided in Appendix~\ref{app:sae_details}.

\subsection{Spatially Structured Visual Sparsity}
\label{sec:spatial}

Token-level sparsity alone does not ensure that a visual feature corresponds to a coherent image region. We therefore adapt the structured visual sparsity construction of \citet{liao2026structured}, grouping image patches using patch--patch attention, spatial proximity, and connectivity-constrained clustering.

For example $i$, let
$\mathcal{G}_i=\{G_{i1},\ldots,G_{iM_i}\}$
denote the partition of image-token positions produced by this clustering procedure, where
$G_{im}\subseteq\mathcal{P}_{\mathrm{img}}$ is region $m$ and
$M_i=|\mathcal{G}_i|$ is the resulting number of regions. Let $q_{itj}\in\{0,1\}$ denote
whether latent $j$ is selected by TopK at position $t$. We keep the TopK selection discrete in the forward pass and use an identity gradient approximation during backpropagation.
Unlike \citealt{liao2026structured}, we 
normalized 
regional support 
by the region size: 
$e_{ijm}
=
\left(
\frac{1}{|G_{im}|}
\sum_{t\in G_{im}} q_{itj}^{2}
\right)^{1/2}.$
Further, the spatial objective for an example with $M_i$ regions is
$\mathcal{L}_{\mathrm{spatial}}^{(i)}
=
\lambda_{\mathrm{group}}
\sum_{j,m} e_{ijm}
+
\frac{\lambda_{\mathrm{exclusive}}}{M_i}
\sum_j
\left(\sum_m e_{ijm}\right)^2,$
where we additionally scale the
exclusivity penalty term by $M_i$. 
These normalizations reduce sensitivity to region size and partition granularity, encouraging each latent's visual support to concentrate on a small number of coherent regions.

\subsection{Phrase--Region Alignment}
\label{sec:alignment}

Spatial structure determines \emph{where} a visual latent activates, but does not ensure that the same latent coordinate captures corresponding linguistic and visual evidence. We therefore introduce a phrase--region alignment objective that grounds shared SAE features across text phrases and image regions.

For example $i$, let $Q_{iq}$ denote the text-token positions in phrase $q$, and let $G_{im}$ denote the image-token positions in region $m$ from \S\ref{sec:spatial}. Given SAE pre-activations
$
[a_{i1}\;\cdots\;a_{iT_i}]^\top\in\mathbb{R}^{T_i\times h},
$
we compute an RMS-pooled representation for any phrase or region $g$:
$s_{ig}
=
\sqrt{
\frac{1}{|g|}
\sum_{t\in g}
a_{it}\odot a_{it}
},
\quad
\bar{s}_{ig}
=
{s_{ig}}/{\|s_{ig}\|_2},$
where the square root is applied element-wise and $s_{ig}\in\mathbb{R}^h$. We pool pre-activations before TopK and ReLU so that phrase--region alignment can use features even when they are not selected at every individual position.

To align phrases with image regions, 
we first define a latent-space similarity between phrase $q$ and image region $m$, as
$z_{iqm}
=
{
\bar{s}_{iQ_{iq}}^\top
\bar{s}_{iG_{im}}
}/{\tau},$
where $\tau>0$ is the temperature.
To supervise this alignment, we must identify which region corresponds to each phrase.
We use the frozen VLM's phrase-to-patch attention as a proxy for this ground truth.
Let
$
A_i\in\mathbb{R}^{Q_i\times M_i}
$
denote the matrix of VLM attention scores (affinities) between the  $Q_i$ phrases and $M_i$ regions in example $i$. For each phrase 
$q$, we select the region with the highest attention score as its positive match:
$\pi_i(q)
=
\arg\max_{m\in\{1,\ldots,M_i\}}
[A_i]_{qm}.$
We then align each phrase with its matched region using a within-image contrastive objective:
$\mathcal{L}_{\mathrm{align}}^{(i)}
=
-\frac{1}{Q_i}
\sum_{q=1}^{Q_i}
\log
\frac{\exp(z_{iq,\pi_i(q)})}
{\sum_{m=1}^{M_i}\exp(z_{iqm})},$
where the remaining regions in the same image serve as negatives.
Finally,
the full cross-modal objective jointly minimizes alignment error, phrase-region magnitude mismatch, and excessive modality imbalance among strongly participating latents:
$\mathcal{L}_{\mathrm{cross}}
=
\lambda_{\mathrm{al}}\mathcal{L}_{\mathrm{align}}
+
\lambda_{\mathrm{mag}}\mathcal{L}_{\mathrm{mag}}
+
\lambda_{\mathrm{bal}}\mathcal{L}_{\mathrm{bal}}.$
Definitions of the magnitude and balance terms are provided in
Appendix~\ref{app:phrase_region_alignment}.

\subsection{Multimodal Influence Decomposition}
\label{sec:multimodal_influence}

Latent influence identifies \emph{which sparse features} carry the effect of a training example, but not where those features are expressed within that example. 
To recover this localization,
in language models, we can compute the attribution of each individual token in the sequence via 
position-wise SAE activations~\citep{yu2026correcting}. In a VLM, the sequence
contains both text and image tokens,
with each image token corresponding to a spatial location on the image patch grid.
By calculating the position-wise attribution 
$A_{itj}
=
-r_{itj}\frac{\partial P_i(R_i)}{\partial r_{itj}},$
we preserve the exact mapping between the training example's influence and its manifestation at each specific word or image patch.

Text-token contributions can therefore be grouped into phrases, while image-token contributions can be visualized directly on their corresponding patches or aggregated over the spatial regions. For phrase $Q_{iq}\subseteq\mathcal{P}_{\mathrm{txt}}$ and image region
$G_{im}\subseteq\mathcal{P}_{\mathrm{img}}$ of training example $i$, we aggregate
$A^{\mathrm{txt}}_{ijq}
=
\sum_{t\in Q_{iq}} A_{itj},
\quad
A^{\mathrm{img}}_{ijm}
=
\sum_{t\in G_{im}} A_{itj}.$
Together with
$A_{ij}
=
\sum_{t=1}^{T_i} A_{itj},$
this organizes a training example's influence by latent feature and the localized linguistic and visual evidence through which it is expressed:
$\text{prediction}
\leftrightarrow
\text{training example}
\leftrightarrow
\text{sparse latent}
\leftrightarrow
\text{phrase / image patch or region}.$
Details on the SAE-mediated forward path, sequence reduction, and patch-grid mapping are provided in Appendix~\ref{app:influence_impl}.

\subsection{Scalable Influence Estimation with EK-FAC}
\label{sec:ekfac}

\begin{wraptable}{r}{0.35\textwidth}
\centering
\caption{Task performance before and after SAE insertion. CIDEr for caption similarity to human references; accuracy is used for chest X-ray classification and VQA.}
\vspace{-2mm}
\label{tab:task_performance_sae}

\small
\setlength{\tabcolsep}{3.5pt}
\renewcommand{\arraystretch}{1.08}

\resizebox{\linewidth}{!}{%
\begin{tabular}{lllccc}
\toprule
Dataset & Model & Metric & Base & FT & +SAE \\
\midrule
VizWiz
& Qwen
& CIDEr
& 0.0738
& 0.7569
& 0.7471 \\

VizWiz
& Gemma
& CIDEr
& 0.0686
& 0.5799
& 0.5793 \\

\midrule
CheXpert
& Qwen
& Acc.
& 0.3320
& 0.4290
& 0.4295 \\

CheXpert
& Gemma
& Acc.
& 0.0040
& 0.4530
& 0.4560 \\

\midrule
GQA
& Qwen
& Acc.
& 0.5943
& 0.6513
& 0.6361 \\

GQA
& Gemma
& Acc.
& 0.4019
& 0.5914
& 0.5867 \\
\bottomrule
\end{tabular}%
}

\vspace{2pt}
\parbox{\linewidth}{\scriptsize
\textit{Note:} Base results for VizWiz and GQA use task-specific prompting.
Without prompting: VizWiz CIDEr is 0.0001 for Qwen and 0.0003 for Gemma,
while GQA accuracy is 0.0001 for Qwen and 0.0000 for Gemma.}

\vspace{-5mm}
\end{wraptable}

The main computational bottleneck is the inverse-curvature product
$
C^{-1}g_i(R_i)
$
in
$
I(z_i\!\rightarrow\!z_\star)
=
-g_\star^\top C^{-1}g_i(R_i).
$ 
Constructing or inverting $C$ directly is infeasible at VLM scale (we write $C=C_{\theta_2}$). We therefore use Eigenvalue-Corrected Kronecker-Factored Approximate Curvature (EK-FAC) \citep{george2018fast,grosse2023studying}, which approximates the inverse curvature layer by layer while retaining structured second-order information.
For a linear layer with input activation $u$ and output gradient $\delta$, K-FAC approximates its curvature block as
$F \approx A\otimes S,
\quad
A=\mathbb{E}[uu^\top],
\quad
S=\mathbb{E}[\delta\delta^\top].$
Writing
$
A=Q_A\Lambda_AQ_A^\top
$
and
$
S=Q_S\Lambda_SQ_S^\top,
$
EK-FAC retains the resulting Kronecker eigenbasis but replaces the factored eigenvalues with corrected values $\Lambda_{\mathrm{EK}}$ estimated from per-example gradients. For a layer gradient reshaped as $V$, the damped inverse-curvature product is approximated by
$C^{-1}\operatorname{vec}(V)
\approx
\operatorname{vec}\!\left[
Q_S
\left(
\frac{Q_S^\top VQ_A}
{\Lambda_{\mathrm{EK}}+\lambda}
\right)
Q_A^\top
\right],$
where division is element-wise and $\lambda>0$ is the damping coefficient.

Thus, rather than constructing the full curvature matrix, EK-FAC rotates each layer gradient into a structured eigenbasis, rescales each direction by its damped inverse curvature, and rotates it back. We apply this operation layerwise to compute the full curvature-corrected influence for each retrieved training example and, through $P_i(R_i)$, its latent decomposition from \S\ref{sec:multimodal_influence}. The curvature statistics are estimated once and reused across training--test pairs. Training examples are ranked by the magnitude of the signed influence estimate,
$|I(z_i\!\rightarrow\!z_\star)|$.
Numerical comparisons with alternative curvature estimators and implementation details are provided in Appendices~\ref{sec:appendix:curvature}, ~\ref{app:candidate_retrieval}, and~\ref{app:influence_impl}.
\section{Experiments}
\label{sec:experiments}

\subsection{Experimental Setup}
\label{sec:exp:setup}

\begin{wrapfigure}{l}{0.4\textwidth}
    \vspace{-5mm}
    \centering
    \includegraphics[width=\linewidth]{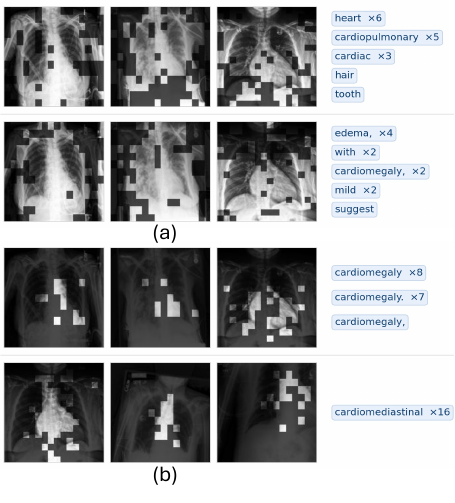}
   \vspace{-6mm}
    \caption{Comparison on CheXpert. Highlighted regions and words indicate latent activations. (a) Base SAE yields diffuse support and weaker radiological alignment. (b) Ours improves localization and clinically relevant alignment.
}
    \label{fig:chexpert_latents}
    \vspace{-10mm}
\end{wrapfigure}

\textbf{Models.}
We evaluate Qwen3-VL-2B-Instruct~\citep{bai2025qwen3vl} and Gemma-4-E2B-it~\citep{gemmateam2026gemma4}, two open-weight VLMs from distinct model families. For each task, we finetune the backbone with LoRA, freeze the VLM, and train the SAE on its task-adapted intermediate representations. We place the SAE at a mid-to-late transformer layer to preserve downstream behavior while exposing semantically meaningful multimodal features. Additional details are provided in Appendix~\ref{app:implementation} and~\ref{sec:app:scale}.

\textbf{Datasets and tasks.}
We evaluate three complementary vision--language settings. \textbf{VizWiz-Captions} tests open-ended captioning under challenging real-world imagery~\citep{gurari2020captioning}. \textbf{CheXpert} tests three-way classification of radiographic findings as \emph{absent}, \emph{uncertain}, or \emph{present}~\citep{irvin2019chexpert}. \textbf{GQA} evaluates compositional visual question answering over objects, attributes, and relations~\citep{hudson2019gqa}.


\subsection{Grounded Multimodal Representation}
\label{sec:exp:representation}

Before using the learned latents for attribution, we verify that the SAE preserves task behavior while providing the spatial and cross-modal structure required for grounded interpretation.

\textbf{Behavior preservation.}
Table~\ref{tab:task_performance_sae} compares each task-adapted VLM before and after SAE insertion. Across all datasets and backbones, performance changes only marginally, indicating that the sparse bottleneck preserves the behavior being explained. Full metrics in Appendix~\ref{sec:appendix:vizwiz_metrics}.

\begin{wrapfigure}{r}{0.45\textwidth}
    \vspace{-2mm}
    \centering
    \includegraphics[width=\linewidth]{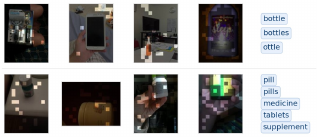}
    \caption{
   Latent breakdown for a VizWiz example with Qwen. The top two latents following Fig.~\ref{fig:influence_chain} capture medicine- and bottle-related evidence; highlighted regions and words indicate representative activations.
    }
    \label{fig:latent_breakdown}
    \vspace{-8mm}
\end{wrapfigure}

\textbf{Evaluation framework.}
We evaluate the learned basis along four dimensions: dictionary health, image--text balance, spatial localization, and semantic coherence. Visual footprint and contiguity measure how concentrated and spatially coherent a latent's visual support is, while MS Lift and causal grounding assess whether that support is semantically aligned with the corresponding linguistic representation. Full metric definitions and additional diagnostics are provided in Appendix~\ref{app:latent_evaluation}.

\textbf{Objective ablation.}
Table~\ref{tab:latent_quality_vizwiz} shows that structured sparsity and phrase--region alignment contribute complementary benefits. Structured sparsity reduces visual footprint from $0.312$ to $0.120$, while phrase--region alignment improves MS Lift from $0.0139$ to $0.0191$. Combining them yields the strongest overall representation, with footprint $0.083$, contiguity $0.548$, balance $26.7\%$, MS Lift $0.0214$, and grounding $0.084$. This suggests that spatial regularization primarily controls \emph{where} a feature activates, while phrase--region alignment improves \emph{what} that localized support represents across modalities. Additional comparisons can be found in Appendix~\ref{app:pooled_sae_baselines}.

\textbf{Generalization across tasks and backbones.}
The same trends hold across datasets and model families (Table~\ref{tab:latent_quality_generalization}). Visual footprint drops from $0.688$ to $0.120$ on GQA with Qwen and from $0.665$ to $0.143$ on VizWiz with Gemma, alongside improvements in dictionary health and modality balance. Fig.~\ref{fig:chexpert_latents} shows the same effect qualitatively on CheXpert, where our latents exhibit sharper visual support and more relevant radiological terminology. Additional qualitative comparisons are provided in Appendix~\ref{app:qualitative_attribution}.

\begin{table*}[t]
\centering

\begin{minipage}[t]{0.48\textwidth}
\centering
\captionof{table}{
Latent-quality ablation on VizWiz with Qwen. Lower footprint and contiguity indicate more localized visual support; higher balance, MS Lift, and grounding indicate stronger multimodal and semantic structure. Grounding significance is assessed by permutation test.
}
\label{tab:latent_quality_vizwiz}

\vspace{-4pt}
\scriptsize
\setlength{\tabcolsep}{2.2pt}
\renewcommand{\arraystretch}{1.08}

\resizebox{\linewidth}{!}{%
\begin{tabular}{lccccccc}
\toprule
Method
& Dead$\downarrow$
& Bal.$\uparrow$
& Foot.$\downarrow$
& Contig.$\downarrow$
& MS$\uparrow$
& Ground.$\uparrow$
& $p$ \\
\midrule

\rowcolor{gray!15}
Base SAE
& $74$
& $22.5$
& $0.312$
& $0.641$
& $0.0139$
& $0.048$
& $9.3\!\times\!10^{-4}$ \\

+ Struct.
& $49$
& $5.5$
& $0.120$
& $0.574$
& $0.0049$
& $0.012$
& $3.1\!\times\!10^{-4}$ \\

+ Align.
& $40$
& $15.7$
& $0.359$
& $0.611$
& $0.0191$
& $0.039$
& $1.2\!\times\!10^{-3}$ \\

Full
& $\mathbf{36}$
& $\mathbf{26.7}$
& $\mathbf{0.083}$
& $\mathbf{0.548}$
& $\mathbf{0.0214}$
& $\mathbf{0.084}$
& $\mathbf{2.9\!\times\!10^{-4}}$ \\

\bottomrule
\end{tabular}%
}
\end{minipage}
\hfill
\begin{minipage}[t]{0.50\textwidth}
\centering
\captionof{table}{
Latent quality across tasks and backbones. Parentheses report the fraction of active latents satisfying the $25\%$ footprint threshold used for MS Lift. Visual MS is omitted for CheXpert.
}
\label{tab:latent_quality_generalization}

\vspace{-4pt}
\scriptsize
\setlength{\tabcolsep}{2.5pt}
\renewcommand{\arraystretch}{1.08}

\resizebox{0.8\linewidth}{!}{%
\begin{tabular}{lllccccc}
\toprule
Dataset & Model & Method
& Dead$\downarrow$
& Bal.$\uparrow$
& Contig.$\downarrow$
& Foot.$\downarrow$
& MS Lift$\uparrow$ \\
\midrule

\rowcolor{gray!12}
VizWiz & Gemma & Base SAE
& $3642$
& $10.1$
& $0.302$
& $0.665$
& $0.0130$ {\tiny ($0.05$)} \\

& & Ours
& $\mathbf{206}$
& $\mathbf{18.2}$
& $\mathbf{0.202}$
& $\mathbf{0.143}$
& $\mathbf{0.0233}$ {\tiny ($\mathbf{0.96}$)} \\

\midrule

\rowcolor{gray!12}
CheXpert & Qwen & Base SAE
& $\mathbf{23}$
& $4.9$
& $0.629$
& $0.577$
& --- \\

& & Ours
& $55$
& $\mathbf{23.9}$
& $\mathbf{0.526}$
& $\mathbf{0.092}$
& --- \\

\rowcolor{gray!12}
& Gemma & Base SAE
& $315$
& $7.3$
& $0.432$
& $0.956$
& --- \\

& & Ours
& $\mathbf{158}$
& $\mathbf{38.7}$
& $\mathbf{0.340}$
& $\mathbf{0.178}$
& --- \\

\midrule

\rowcolor{gray!12}
GQA & Qwen & Base SAE
& $24$
& $12.0$
& $0.603$
& $0.688$
& $0.0309$ {\tiny ($0.20$)} \\

& & Ours
& $\mathbf{7}$
& $\mathbf{13.2}$
& $\mathbf{0.481}$
& $\mathbf{0.120}$
& $\mathbf{0.0394}$ {\tiny ($\mathbf{1.00}$)} \\

\rowcolor{gray!12}
& Gemma & Base SAE
& $5759$
& $3.3$
& $0.652$
& $0.769$
& --- {\tiny ($0.00$)} \\

& & Ours
& $\mathbf{184}$
& $\mathbf{9.4}$
& $\mathbf{0.261}$
& $\mathbf{0.246}$
& $\mathbf{0.0255}$ {\tiny ($\mathbf{0.56}$)} \\

\bottomrule
\end{tabular}%
}
\end{minipage}

\vspace{-3mm}
\end{table*}

\subsection{Validating Latent Influence: Necessity and Sufficiency}
\label{sec:exp:causal}

Grounded features are useful for attribution only if they actually matter to the prediction. We therefore test whether high-influence latents are \emph{necessary} for producing the output and \emph{sufficient} to recover it.

For each test example, we aggregate influence across retrieved training examples and rank latents by this score. Necessity removes the top-$k$ latents, while sufficiency keeps only the top-$k$. We compare against activation magnitude, firing frequency, and random rankings for $k\in[64,1024]$. Necessity is evaluated by changes in correct-token logit, NLL, and prediction flips; sufficiency by same-answer rate and retained correct-token logit. Implementation details are in Appendix~\ref{app:influence_impl}.

\begin{figure}[t]
\vspace{-2mm}
    \centering
    \includegraphics[width=0.8\linewidth]{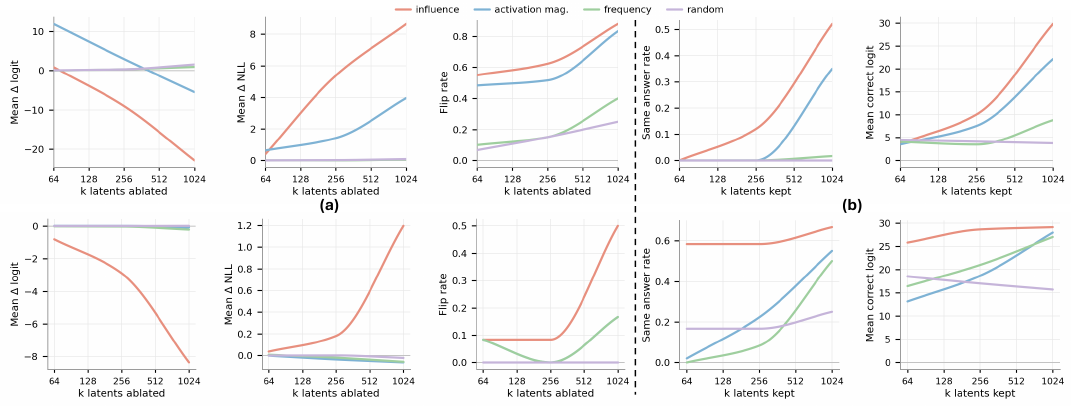}
    \vspace{-2mm}
    \caption{
CheXpert necessity--sufficiency. Qwen [top], Gemma [bottom]; (a) necessity, (b) sufficiency.
    }
    \label{fig:necsuff_chexpert}
    \vspace{-2mm}
\end{figure}

\begin{figure}[t]
    \centering
    \includegraphics[width=0.8\linewidth]{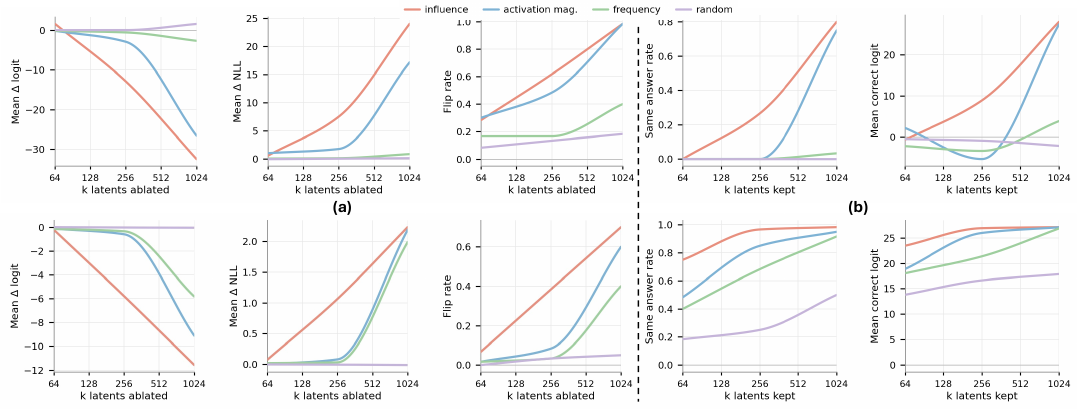}
    \vspace{-2mm}
    \caption{
GQA necessity--sufficiency. Qwen [top], Gemma [bottom]; (a) necessity, (b) sufficiency.
    }
    \label{fig:necsuff_gqa}
    \vspace{-5mm}
\end{figure}

Figs.~\ref{fig:necsuff_vizwiz}--\ref{fig:necsuff_gqa} show consistent separation across datasets and backbones. Removing influence-ranked latents generally causes substantially larger prediction changes than removing activation-, frequency-, or randomly ranked features. For example, on GQA with Gemma at $k=256$, influence-based removal flips $38.3\%$ of predictions, compared with $8.3\%$ under activation ranking. Similarly, the complementary sufficiency results show that the same features preserve substantially more of the original prediction when isolated. On GQA with Gemma, only $64$ influence-ranked latents recover the original answer for $75.0\%$ of examples, compared with $48.3\%$ for activation magnitude. Similar trends hold on VizWiz and CheXpert. Thus, the latents surfaced by the influence estimator are not merely descriptive features: they carry important information for the model's prediction.

\label{sec:exp:necsuff}
\begin{figure}[t]
    \centering
    \includegraphics[width=0.8\linewidth]{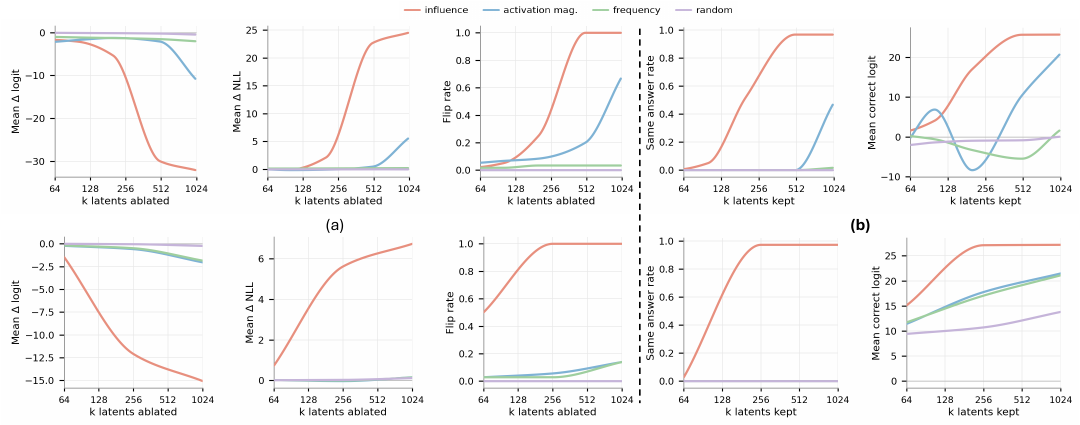}
    \vspace{-2mm}
    \caption{
VizWiz necessity--sufficiency. Qwen [top], Gemma [bottom]; (a) necessity, (b) sufficiency.
    }
    \label{fig:necsuff_vizwiz}
    \vspace{-2mm}
\end{figure}

\subsection{How Visual Evidence Shapes Training-Data Influence}
\label{sec:exp:visual_evidence}

\begin{wrapfigure}{r}{0.3\textwidth}
    \vspace{-10mm}
    \centering
    \includegraphics[width=\linewidth]{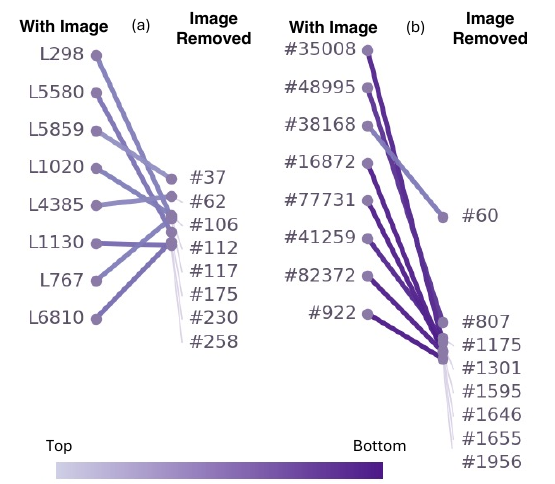}
    \vspace{-8pt}
    \caption{
    Effect of removing visual evidence on VizWiz influence rankings with Qwen. (a) Latent rankings remain relatively stable. (b) Training-example rankings change substantially.
    }
    \label{fig:visual_evidence_influence}
    \vspace{-8mm}
\end{wrapfigure}

The shared image--text latent space lets us ask whether removing visual evidence changes \emph{which training examples} influence a prediction more strongly than it changes the \emph{concepts} through which that influence is expressed. We evaluate this by recomputing influence for the same test examples with the image removed, while keeping the SAE and influence-estimation procedure fixed.

Across six model--dataset settings, absolute latent influence remains moderately to strongly correlated between the multimodal and text-only conditions, with Spearman $\rho$ ranging from $0.543$ to $0.795$. In contrast, training-example attribution is substantially reorganized, with example-level correlations ranging from $-0.797$ to $-0.564$. Figure~\ref{fig:visual_evidence_influence} shows a representative example, where removing the image sharply reorders the influential training examples but maintains latent organization.

These results suggest that visual evidence primarily changes the \emph{training-data support} underlying a prediction, while the latent concepts carrying influence remain comparatively more stable. The learned multimodal latent space provides a relatively stable conceptual interface, whereas the training examples associated with those concepts remain strongly conditioned on the available visual evidence. Additional analyses and results are provided in Appendix~\ref{app:modality_analysis}.

\subsection{Hierarchical Multimodal Attribution}
\label{sec:exp:qual}
\begin{wrapfigure}{r}{0.5\textwidth}
    \vspace{-2mm}
    \centering
    \includegraphics[width=\linewidth]{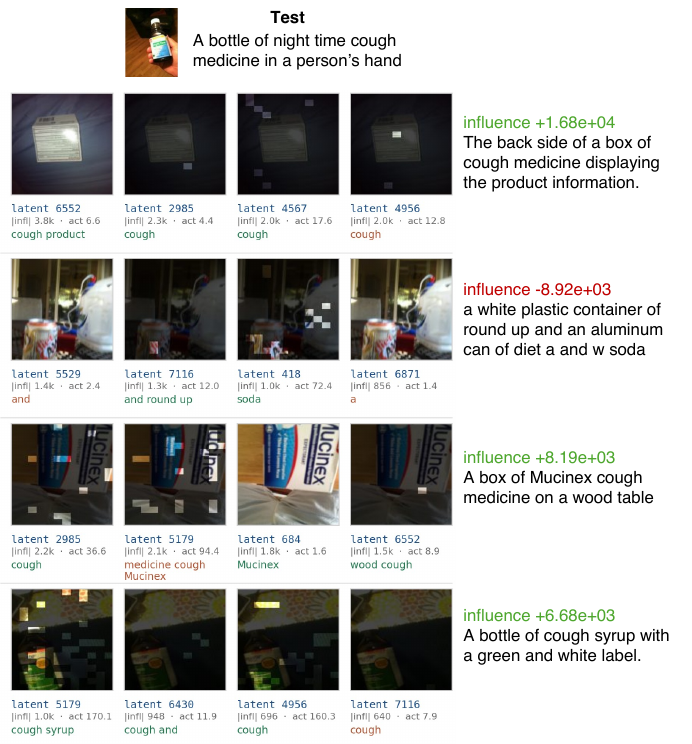}
    \caption{
Training-example breakdown for a VizWiz example with Qwen. Training examples are ranked by influence; we show the top contributing latents, with peak image activations and caption words. Additional results are in the appendix.
    }
    \label{fig:influence_chain}
    \vspace{-5mm}
\end{wrapfigure}

The preceding experiments establish that the learned representation is grounded and that influence-ranked latents are behaviorally meaningful. We now examine what this decomposition reveals about individual predictions.

Fig.~\ref{fig:influence_chain} illustrates the full attribution hierarchy for a representative VizWiz example. Influential training examples contain prediction-relevant concepts such as cough medicine and related containers, while decomposing their influence through individual latents reveals finer visual and linguistic evidence, including medicine, bottle, and the wooden surface on which the object appears. 
The same latent contributions can be localized to both text phrases and image regions, exposing not only which training examples matter but also the grounded concepts through which their influence is expressed.

Fig.~\ref{fig:latent_breakdown} provides the complementary test-conditioned view, directly ranking influential latents on the query representation. Here, the highest-ranked features capture medicine- and bottle-related evidence localized to the corresponding parts of the image. 
Together, the two views separate \emph{where influence originates} from \emph{what multimodal evidence carries it}.
Additional test-latent decompositions and influential-training-example analyses are provided in Appendix~\ref{app:qualitative_attribution}.

\begin{figure}[t]
    \centering
    \includegraphics[width=0.75\linewidth]{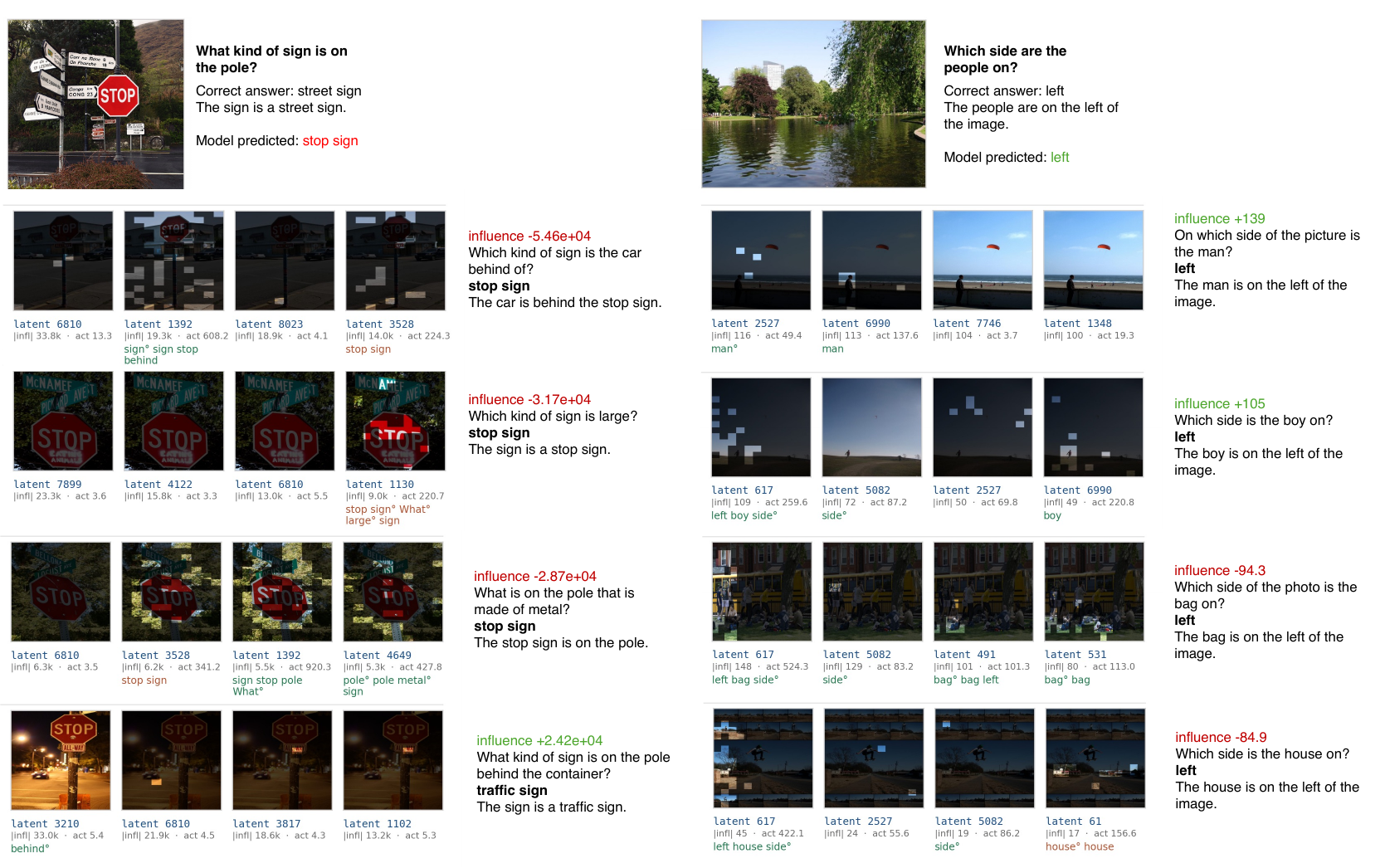}
    \vspace{-2mm}
    \caption{
GQA influence decompositions. Incorrect prediction [left]. Correct prediction [right].
}
    \label{fig:gqa_correct_incorrect}
    \vspace{-5mm}
\end{figure}

\subsection{Attribution for Correct and Incorrect Predictions}
\label{sec:exp:correct_incorrect}

We further examine whether the attribution hierarchy helps distinguish successful from failed predictions. Fig.~\ref{fig:gqa_correct_incorrect} contrasts two representative GQA examples: one answered correctly by the model and one answered incorrectly. For each, we trace the prediction to its most influential training examples and decompose their influence through the highest-contributing multimodal latents.

In the correct case, the influential examples and latent concepts align with the relevant visual evidence and support the predicted answer. In the incorrect case, they instead concentrate on visually salient evidence associated with the model's mistaken prediction, revealing how the model can rely on the wrong grounded concept despite attending to relevant image content. This comparison shows how grounded influence can diagnose not only \emph{what} supports a prediction, but also \emph{why} that support matters.

\section{Conclusion and Discussion}
\label{sec:conclusion}

We introduced \method{}, a framework for tracing VLM predictions from influential training examples
through sparse multimodal concepts to localized linguistic and visual evidence. By combining a shared
image--text latent space with spatially coherent visual structure, phrase--region alignment, and
latent-mediated influence, \method{} provides a grounded view of how training data shapes individual
predictions. Across two VLM families and three tasks, the learned representation preserves model
behavior while yielding more localized and semantically coherent features, and influence-ranked
latents are substantially more behaviorally consequential than activation- or frequency-based
alternatives. Our analysis further shows that visual evidence can strongly reorganize which training
examples matter while preserving substantially more of the underlying latent structure. These results
provide a path toward more fine-grained and auditable training-data attribution in multimodal models. Limitations and future work are discussed in Appendix~\ref{sec:limitations}.
\section*{Acknowledgements}
This work was supported in part by National Institutes of Health (NIH) grants R01EB037101 and R01MH143267, as well as AWS Promotional Credits provided through the Amazon Web Services (AWS) Health Equity Initiative. The content is solely the responsibility of the authors and does not necessarily represent the official views of the NIH.

\newpage
\bibliography{999_reference}
\bibliographystyle{style/icml2025}

\titlespacing*{\section}{0pt}{*1}{*1}
\titlespacing*{\subsection}{0pt}{*1.25}{*1.25}
\titlespacing*{\subsubsection}{0pt}{*1.5}{*1.5}

\setlength{\abovedisplayskip}{\baselineskip} 
\setlength{\abovedisplayshortskip}{0.5\baselineskip} 
\setlength{\belowdisplayskip}{\baselineskip}
\setlength{\belowdisplayshortskip}{0.5\baselineskip}

\clearpage
\appendix
\label{sec:append}
\part*{Appendix}
{
\setlength{\parskip}{-0em}
\startcontents[sections]
\printcontents[sections]{ }{1}{}
}

\setlength{\parskip}{.5em}
\section{Limitations and Future Work}
\label{sec:limitations}

Our approach depends on the quality of the learned SAE representation: latent features may remain partially entangled, and attribution can depend on the chosen layer and dictionary capacity. Phrase--region alignment also relies on pretrained attention as a correspondence signal, which may be noisy or incomplete. In addition, EK-FAC and latent-mediated influence remain approximations to the effect of full retraining. Future work could study multi-layer or hierarchical latent spaces, attention-independent grounding objectives, and stronger validation against retraining-based influence. Extending grounded latent influence to larger VLMs, video and other modalities, and applications such as data debugging, hallucination analysis, and model editing is also promising.

\section{Curvature Approximation Study}
\label{sec:appendix:curvature}

Computing the representation-mediated influence requires applying an inverse-curvature operator to
the test gradient. Explicitly forming this matrix is infeasible even for the upper-model parameter
subset used for influence computation, which contains approximately $6\times10^{8}$ parameters. We therefore evaluated several scalable curvature
approximations before adopting EK-FAC. The alternatives exhibit
qualitatively different failure modes, ranging from unstable linear
solves to curvature estimates that are numerically well behaved but
provide little meaningful geometric correction.

\paragraph{Raw Hessian.}
We first considered implicit Hessian--vector products combined with
conjugate gradients (CG). This approach is unreliable in our setting
because the empirical Hessian is strongly indefinite, whereas standard
CG assumes a positive-definite system. The solver repeatedly encounters
negative curvature and terminates prematurely.

This instability translates directly into unstable influence estimates.
Across
$\texttt{cg\_iters}\in\{10,40,160\}$, the maximum absolute influence
changes from
\[
0.19 \;\rightarrow\; 0.35 \;\rightarrow\; 3.31,
\]
and the solutions obtained with $10$ and $160$ iterations share no
examples among their respective top-$10$ training examples. Thus, the
resulting attribution depends strongly on the optimization trajectory
rather than defining a stable inverse-Hessian estimate.

\paragraph{Generalized Gauss--Newton.}
The generalized Gauss--Newton (GGN) matrix avoids the negative-curvature
issue because it is positive semidefinite by construction. However, its
linear systems remain difficult to solve in our setting. Even under
substantial damping, residuals fail to decrease reliably, while each
solve is substantially more expensive than the alternatives considered
here. Thus, although GGN provides a more
well-behaved curvature object than the raw Hessian, it does not offer a
practical inverse-curvature estimator at the scale required for our
experiments.

\paragraph{Empirical Fisher.}
The empirical Fisher is attractive because it is positive semidefinite
and admits an efficient low-rank inverse. For a curvature cohort of
$m$ examples,
\begin{equation}
F
=
\frac{1}{m}
\sum_{i=1}^{m}
g_i g_i^{\top}.
\label{eq:empirical_fisher}
\end{equation}
With damping, the resulting system is numerically stable and can be
solved efficiently through its low-rank representation. Its main
limitation in our setting, however, is not numerical but geometric.

In particular, the cohort gradients are nearly orthogonal: the mean
pairwise cosine similarity is $-0.002$. As a result, each additional
example contributes an almost independent direction rather than
reinforcing a shared low-dimensional curvature structure. With only
$m=32$ curvature examples in a parameter space of approximately
$6\times10^{8}$ dimensions, the empirical Fisher therefore covers an
extremely small fraction of the directions relevant to the test
gradient. Empirically, doubling the cohort size increases the captured
gradient mass by only approximately $1.3\times$, indicating that this
coverage improves slowly with additional curvature examples.

This low-rank geometry makes the damped inverse close to a scaled
identity outside the cohort span. We quantify this using the alignment
between the inverse-curvature product and the original test gradient,
\begin{equation}
\mathrm{CurvAlign}
=
\cos\!\left(
(F+\lambda I)^{-1}g_{\mathrm{test}},
g_{\mathrm{test}}
\right).
\label{eq:curvature_alignment}
\end{equation}
For the empirical Fisher, we obtain
$\mathrm{CurvAlign}=0.987$, showing that the curvature transformation
leaves the test-gradient direction almost unchanged. In this regime,
the resulting influence score approaches a rescaled first-order
gradient-similarity score rather than substantially incorporating
second-order geometry.

Notably, standard numerical checks do not expose this failure mode.
The Woodbury solve attains residual ratios of approximately $10^{-6}$,
the associated Gram matrix has condition number $2.55$, and influence
rankings remain highly stable across disjoint curvature cohorts
(Spearman $\rho=0.993$). These diagnostics confirm that the low-rank
system is solved accurately, but they do not establish that the
curvature approximation is informative. Indeed, the high
$\mathrm{CurvAlign}$ value shows that the empirical Fisher is stable
largely because most directions remain dominated by the damping term.

Thus, in our setting, the empirical Fisher is not rejected because of
poor conditioning or solver error. Rather, the near-orthogonality of
per-example gradients makes a small-cohort empirical Fisher too
low-rank to induce a meaningful transformation of the test gradient.

\paragraph{EK-FAC.}
We use EK-FAC for the final influence computation; the approximation
itself is described in the main text. EK-FAC provides a structured
layerwise curvature approximation while avoiding the severe rank
restriction of cohort-based curvature estimates. This is particularly
important in our setting, where per-example gradients are nearly
orthogonal and a small collection of training gradients spans only a
negligible portion of the full parameter space. By instead modeling
curvature through layerwise activation and gradient statistics, EK-FAC
captures curvature structure beyond the span of individual examples
while remaining computationally tractable at model scale.

We additionally validate the EK-FAC approximation in controlled
small-scale settings where the target curvature can be explicitly
materialized. When constructed at the example level, consistent with our
influence formulation, EK-FAC achieves cosine fidelity of
$0.997$--$0.9996$ across dense, ragged, and masked conditions. This high
agreement indicates that the structured approximation closely preserves
the curvature transformation relevant to our influence estimates while
remaining tractable at model scale.

These results support EK-FAC as a practical approximation for our
setting: it provides a scalable, non-degenerate curvature transformation
while preserving the example-level geometry required by our influence
formulation. We therefore use EK-FAC for all influence results reported
in the main paper.

\section{Full VizWiz Captioning Results}
\label{sec:appendix:vizwiz_metrics}

Table~\ref{tab:vizwiz_full_metrics} reports the complete set of captioning metrics for VizWiz. LoRA finetuning substantially improves performance over both prompted and unprompted base models, while SAE insertion results in only minor changes across metrics, indicating strong preservation of task-relevant information.

\begin{table*}[t]
\centering
\caption{Full VizWiz captioning results. We report CIDEr, METEOR, BLEU-1--4, and ROUGE-L. ``Prompted'' uses the task-specific captioning prompt, while ``zero-shot'' denotes evaluation without task-specific prompting.}
\label{tab:vizwiz_full_metrics}
\small
\setlength{\tabcolsep}{4.5pt}
\renewcommand{\arraystretch}{1.10}
\begin{tabular}{llccccccc}
\toprule
Model & Setting
& CIDEr$\uparrow$
& METEOR$\uparrow$
& B-1$\uparrow$
& B-2$\uparrow$
& B-3$\uparrow$
& B-4$\uparrow$
& R-L$\uparrow$ \\
\midrule

Qwen
& Finetuned, no SAE
& 0.7569
& 0.2641
& 0.7076
& 0.5171
& 0.3736
& 0.2660
& 0.4767 \\

Qwen
& + SAE
& 0.7471
& 0.2612
& 0.7099
& 0.5180
& 0.3737
& 0.2659
& 0.4758 \\

Qwen
& Base (prompted)
& 0.0738
& 0.2385
& 0.3452
& 0.2070
& 0.1205
& 0.0706
& 0.3079 \\

Qwen
& Base (zero-shot)
& 0.0001
& 0.0026
& 0.0001
& 0.0000
& 0.0000
& 0.0000
& 0.0108 \\

\midrule

Gemma
& Finetuned, no SAE
& 0.5799
& 0.2405
& 0.6608
& 0.4558
& 0.3068
& 0.2030
& 0.4419 \\

Gemma
& + SAE
& 0.5793
& 0.2389
& 0.6622
& 0.4557
& 0.3057
& 0.2016
& 0.4416 \\

Gemma
& Base (prompted)
& 0.0686
& 0.1747
& 0.2902
& 0.1564
& 0.0826
& 0.0449
& 0.2489 \\

Gemma
& Base (zero-shot)
& 0.0003
& 0.0156
& 0.0230
& 0.0010
& 0.0000
& 0.0000
& 0.0265 \\

\bottomrule
\end{tabular}
\end{table*}

\section{Additional Representation Evaluation}
\label{app:latent_evaluation}

We provide additional details for the latent-quality metrics used in
Sec.~\ref{sec:exp:representation}, including dictionary health, modality balance,
visual footprint, contiguity, MS Lift, and causal grounding. We additionally
report diagnostics of latent activity correlation and layer placement to verify
that the selected SAE interface provides a well-separated representation while
preserving downstream behavior.

\subsection{Latent Quality Metrics}
\label{sec:appendix:latent_metrics}

Here we provide precise definitions that
are useful for interpreting these quantities but are omitted from the main
tables for compactness. All statistics are computed on a fixed held-out set of
1,500 training examples, independently of the influence pipeline.

\paragraph{Dead latents.}
A latent is considered dead if it never activates on any of the 1,500
evaluation examples.

\paragraph{Modality balance.}
For latent $j$ and modality $m\in\{\mathrm{img},\mathrm{txt}\}$, we define
the token-normalized activation rate
\begin{equation}
r_j^m =
\frac{\mathrm{mass}_j^m}{N_{\mathrm{tokens}}^m},
\end{equation}
and image share
\begin{equation}
s_j =
\frac{r_j^{\mathrm{img}}}
     {r_j^{\mathrm{img}}+r_j^{\mathrm{txt}}}.
\end{equation}
A live latent is considered balanced when $s_j\in[0.35,0.65]$. Token
normalization prevents modality share from being dominated by differences in
image and text sequence length.

\paragraph{Spatial footprint.}
Spatial footprint measures the fraction of an individual image's patch grid occupied by
the latent.

\paragraph{Spatial contiguity.}
Raw connected-component counts depend strongly on support size, so we normalize
each latent's component count by that of a size-matched random mask on the same
patch grid,
\begin{equation}
C_{\mathrm{rel}}
=
\frac{C(\mathcal{S})}
{\mathbb{E}_{\widetilde{\mathcal{S}}:
|\widetilde{\mathcal{S}}|=|\mathcal{S}|}
[C(\widetilde{\mathcal{S}})]}.
\end{equation}
Values below $1$ indicate support that is more spatially contiguous than
expected by chance. We treat contiguity as supplementary to footprint rather
than as a standalone localization measure.

\paragraph{Visual monosemanticity.}
Following the MonoSemanticity Score (MS) of \citet{pach2025sparse}, we assess
whether the regions activating a latent are semantically coherent under an
independent reference encoder. We adapt MS to localized VLM features by
evaluating the image patches on which the latent itself fires using DINOv2.

Raw MS is strongly confounded by footprint size. We therefore compare the
observed support $\mathcal{S}_j$ with a random support of identical size drawn
from the same images and report
\begin{equation}
\mathrm{MSLift}_j
=
\mathrm{MS}(\mathcal{S}_j)
-
\mathbb{E}\!\left[
\mathrm{MS}(\widetilde{\mathcal{S}}_j)
\right].
\end{equation}
The main text reports this null-corrected lift rather than raw MS.
As the support approaches full-image coverage, the matched random support increasingly overlaps
the observed support and the comparison becomes degenerate. We therefore score only latents whose
median footprint covers less than 25\% of the patch grid.

\paragraph{Why visual MS is omitted for CheXpert.}
For CheXpert, RAD-DINO~\citep{perezgarcia2024raddino} produces a substantially
higher size-matched similarity baseline than DINOv2 does on natural images.
Because chest radiographs exhibit highly stereotyped global anatomy, arbitrary
regions can already be strongly similar across examples. The assumption that a
meaningful latent region should be more self-consistent than its surroundings
is therefore poorly matched to this domain, and we omit visual MS from the
CheXpert evaluation.

\paragraph{Causal grounding.}
For each latent $j$, we ablate it only at image patches where it is active
and measure the change in negative log-likelihood of its top-associated
word. We define
\[
\operatorname{share}(a)
=
\frac{
\operatorname{NLL}(a)-\operatorname{NLL}_{\mathrm{full}}
}{
\operatorname{NLL}_{\mathrm{no\text{-}image}}
-
\operatorname{NLL}_{\mathrm{full}}
},
\]
where $\operatorname{NLL}_{\mathrm{no\text{-}image}}$ is obtained by
zeroing all SAE latents at all image positions. The causal grounding
effect is
\[
\operatorname{GroundingEffect}_j
=
\operatorname{share}(j)
-
\operatorname{share}(c(j)),
\]
where $c(j)$ is a control latent with similar active-patch count, ablated
at its own active patches in the same image. We report the mean effect as
\texttt{mean\_solo\_effect} together with
\texttt{solo\_effect\_pvalue}. We exclude examples with image dependence
below $0.05$ and latents active on fewer than two image patches.

\subsection{Latent-Space Orthogonality}
\label{sec:app:ortho}

Latent-mediated influence is easiest to interpret when individual SAE coordinates are weakly correlated, so that attribution to one latent is not simply a rotated description of a highly redundant subspace. This motivation is closely related to the approximately orthogonal latent basis used for influence attribution by \citet{yu2026correcting}. We therefore evaluate whether our learned sparse representation becomes more orthogonal than the dense VLM activations from which it is constructed.

For Qwen, we measure orthogonality at three representation levels: the dense activations entering the SAE (\emph{pre-latent}), the $k$-sparse SAE code (\emph{SAE latent}), and the input text embeddings (\emph{text}) as a reference. For each representation, we center and normalize the features, form the corresponding Gram matrix, and report the mean absolute and mean squared off-diagonal entries, the fraction of near-orthogonal pairs satisfying $|\rho|<0.1$, and the stable rank
\begin{equation}
\mathrm{srank}(A)
=
\frac{\|A\|_F^2}{\|A\|_2^2}.
\end{equation}
Table~\ref{tab:ortho} reports results for our full SAE at layer $\ell=15$.

The SAE substantially decorrelates the representation on both datasets. On VizWiz, the fraction of near-orthogonal pairs increases from $20.8\%$ in the dense representation to $99.5\%$ after SAE encoding, while the mean absolute off-diagonal correlation decreases from $0.280$ to $0.0020$. CheXpert shows the same pattern, with near-orthogonal pairs increasing from $25.4\%$ to $99.2\%$ and mean absolute correlation decreasing from $0.239$ to $0.0036$. Thus, despite the different visual domains, the learned sparse codes consistently provide a substantially less correlated coordinate system for subsequent per-latent influence analysis.

\begin{table}[t]
\centering
\caption{Orthogonality of Qwen representations before and after SAE encoding at layer $\ell=15$ ($d_{\mathrm{latent}}=8192$, $k=128$).}
\label{tab:ortho}
\small
\setlength{\tabcolsep}{4pt}
\renewcommand{\arraystretch}{1.08}
\begin{tabular}{llccc}
\toprule
Dataset & Metric & Text & Pre-latent & SAE latent \\
\midrule

\multirow{4}{*}{VizWiz}
& Near-orthogonal ($|\rho|<0.1$) $\uparrow$
& $7.6\%$ & $20.8\%$ & $\mathbf{99.5\%}$ \\
& Mean $|\mathrm{offdiag}(G)|$ $\downarrow$
& $0.798$ & $0.280$ & $\mathbf{0.0020}$ \\
& Mean $\mathrm{offdiag}(G)^2$ $\downarrow$
& $0.751$ & $0.115$ & $\mathbf{0.0008}$ \\
& Stable rank (raw / norm.)
& $1.16/0.0053$ & $4.01/0.0020$ & $9.93/0.0012$ \\

\midrule

\multirow{4}{*}{CheXpert}
& Near-orthogonal ($|\rho|<0.1$) $\uparrow$
& $24.9\%$ & $25.4\%$ & $\mathbf{99.2\%}$ \\
& Mean $|\mathrm{offdiag}(G)|$ $\downarrow$
& $0.510$ & $0.239$ & $\mathbf{0.0036}$ \\
& Mean $\mathrm{offdiag}(G)^2$ $\downarrow$
& $0.424$ & $0.086$ & $\mathbf{0.0010}$ \\
& Stable rank (raw / norm.)
& $1.54/0.0044$ & $4.71/0.0023$ & $17.25/0.0021$ \\

\bottomrule
\end{tabular}
\end{table}

\paragraph{Interpretation.}
The near-orthogonal fraction and off-diagonal Gram statistics provide the most direct comparisons because they measure correlation between coordinates and are less sensitive to representation width. The raw stable rank should be interpreted more cautiously: the SAE latent space is $8192$-dimensional, compared with $2048$ dimensions for the pre-latent representation, so its absolute value is not directly comparable across spaces. Indeed, after normalization by the maximum attainable rank, the ordering can change. We therefore use stable rank only as a supplementary diagnostic and base our orthogonality claim on the near-orthogonal fraction and off-diagonal correlations.

The text representation is likewise included only as a reference. Its Gram statistics are computed over the token axis rather than the SAE feature axis and therefore should not be interpreted as a direct intermediate point between the dense and sparse representations. Our main conclusion is restricted to the comparable pre-latent and SAE-latent representations: SAE encoding substantially reduces coordinate correlation, providing a near-orthogonal basis better suited to latent-mediated influence attribution.

\subsection{Sensitivity to SAE Layer Placement}
\label{sec:appendix:layer_sensitivity}

We evaluate SAE placement sensitivity on Gemma for GQA using the same 8192-dimensional SAE with $k=128$. While the primary layer-19 placement preserves fine-tuned performance, moving the SAE to layer 12 reduces accuracy to $0.3315$.

\begin{table}[t]
\centering
\small
\caption{Sensitivity of Gemma on GQA to SAE placement.}
\label{tab:gemma_layer_sensitivity}
\begin{tabular}{lcc}
\toprule
\textbf{Setting} & \textbf{SAE Layer} & \textbf{GQA Acc.} \\
\midrule
Base model & -- & 0.4019 \\
Fine-tuned model & -- & 0.5914 \\
Fine-tuned + SAE & 19 & 0.5867 \\
Fine-tuned + SAE & 12 & 0.3315 \\
\bottomrule
\end{tabular}
\end{table}

This approximately 26-point drop shows that SAE placement materially affects representation fidelity as shown in Table~\ref{tab:gemma_layer_sensitivity}, dropping below base performance. We therefore use layer 19 for the main Gemma experiments, where the SAE preserves downstream behavior while providing the latent basis used for attribution.

\section{Gradient-Based Candidate Retrieval}
\label{app:candidate_retrieval}

Computing full latent-mediated influence for every training--test pair is costly.
Following \citet{yu2026correcting}, we first retrieve candidate training examples
using gradient similarity and apply the full curvature-aware influence estimator
only to the retained set. For test example $z_{\mathrm{test}}$ and training example
$z_i$, we compute
$
s_i
=
\left\langle
\nabla_\theta \mathcal{L}(z_i;\theta),
\nabla_\theta \mathcal{L}(z_{\mathrm{test}};\theta)
\right\rangle
$
and retain
$
\mathcal{C}_K(z_{\mathrm{test}})
=
\operatorname{TopK}_i\,s_i.
$
Gradient similarity is used only for candidate retrieval; all reported training-example rankings are determined by the latent-mediated EK-FAC influence estimator.

\section{Computational Scope and Model Scale}
\label{sec:app:scale}

All SAE training and influence analyses are run on a single NVIDIA RTX PRO Blackwell GPU. At this setup, the approximately 2B-parameter models used in our experiments are already near the practical memory and compute limit for the full pipeline, which requires per-example gradients, EK-FAC statistics, and latent-resolved influence computation. We therefore restrict evaluation to this scale and leave larger VLMs to future work.

\section{Latent-Mediated Influence Implementation}
\label{app:influence_impl}

\paragraph{SAE-mediated forward path.}
The SAE is inserted directly into the VLM forward computation at the selected
intermediate layer. The upper model receives the reconstructed representation
\[
\hat X_i^l = W_{\mathrm{dec}}R_i + b_{\mathrm{pre}},
\]
rather than the original residual stream. Thus, all influence computations,
latent interventions, and downstream evaluations are performed through the same
SAE-mediated representation used for attribution.

\paragraph{Sequence reduction.}
For multimodal example $i$, the SAE representation is sequence-resolved,
$
R_i\in\mathbb{R}^{T_i\times h}.
$
For the global training-example influence computation, we reduce derivatives
over valid sequence positions to obtain a common $h$-dimensional latent
representation. Token-resolved activations are retained for the subsequent
phrase- and region-level localization in Sec.~\ref{sec:multimodal_influence}. This reduction provides a consistent latent representation across
variable-length image--text sequences for curvature and example-level influence,
while the original sequence-resolved activations and derivatives are retained
for phrase-, patch-, and region-level localization.

\paragraph{Position-resolved multimodal localization.}
Although example-level influence is computed using EK-FAC in model-parameter space,
we decompose this influence through the SAE latent representation. For the global
latent decomposition, derivatives are aggregated over valid sequence positions to
obtain an $h$-dimensional latent representation, while the sequence-resolved SAE
activations $R_i\in\mathbb{R}^{T_i\times h}$ and their position-wise derivatives are retained
for localization. Thus, for latent $j$, the contribution
\[
A_{itj}
=
-r_{itj}\frac{\partial P_i}{\partial r_{itj}}
\]
remains indexed by the original sequence position $t$.

For text positions, $t$ maps directly to the corresponding token and can be
aggregated over phrase spans $Q_q$. For image positions, we retain the VLM's
mapping from each visual token to its location on the image patch grid. We
therefore reshape $\{A_{itj}:t\in\mathcal{P}_{\mathrm{img}}\}$ back onto that
grid to obtain a patch-level attribution map for latent $j$. Region-level
attribution is obtained by summing the patch contributions belonging to region
$G_m$:
\[
A^{\mathrm{img}}_{ijm}
=
\sum_{t\in G_m} A_{itj},
\qquad
A^{\mathrm{txt}}_{ijq}
=
\sum_{t\in Q_q} A_{itj}.
\]
The patch-level maps are used for visualization, while phrase- and region-level
aggregates provide the grouped multimodal decomposition reported in the main
text.

\paragraph{Candidate retrieval.}
We use the gradient-based filtering procedure described in
Appendix~\ref{app:candidate_retrieval} before evaluating full EK-FAC influence.

\subsection{SAE and Decorrelation Details}
\label{app:sae_details}

For completeness, we summarize the full representation-learning objective used
to train the multimodal SAE. For a minibatch $\mathcal{B}$, the objective is
\[
\mathcal{L}_{\mathrm{SAE}}
=
\mathcal{L}_{\mathrm{rec}}
+
\lambda_{\perp}\mathcal{L}_{\perp}
+
\lambda_{\mathrm{aux}}\mathcal{L}_{\mathrm{aux}}
+
\lambda_{\mathrm{lb}}\mathcal{L}_{\mathrm{lb}}
+
\lambda_{\mathrm{group}}\mathcal{L}_{\mathrm{group}}
+
\lambda_{\mathrm{exclusive}}\mathcal{L}_{\mathrm{exclusive}}
+
\mathcal{L}_{\mathrm{cross}}.
\]

The load-balancing term
$\mathcal{L}_{\mathrm{lb}}$ discourages activity from collapsing onto a small
subset of the dictionary.

\paragraph{Reconstruction objective.}
The SAE is trained with
\[
\mathcal{L}_{\mathrm{rec}}
=
\frac{1}{T}
\sum_{t=1}^{T}
\|x_t^l-\hat{x}_t^l\|_2^2.
\]
The VLM remains frozen throughout SAE training.

\paragraph{Decorrelated latent activity.}
For latent $j$, let
$
r_j=[r_{1j}\;\cdots\;r_{Tj}]^\top
$
and define its centered, normalized activity as
\[
\tilde r_j
=
\frac{r_j-\bar r_j\mathbf{1}}
{\|r_j-\bar r_j\mathbf{1}\|_2+\epsilon}.
\]
We penalize off-diagonal correlations,
\[
\mathcal{L}_{\perp}
=
\frac{1}{h(h-1)}
\sum_{i\neq j}
(\tilde r_i^\top\tilde r_j)^2.
\]
This encourages approximately decorrelated latent activity without imposing
orthogonality on the decoder dictionary.

\subsubsection{Additional Phrase--Region Alignment Details}
\label{app:phrase_region_alignment}

Section~\ref{sec:alignment} introduces the phrase--region alignment objective used to ground shared SAE features across image and text. Here we provide the additional magnitude- and modality-balance terms used in the full cross-modal objective.

\paragraph{Response-magnitude matching.}
Contrastive alignment constrains the direction of normalized phrase and region representations but does not control their relative activation scale. For example $b$, we define the ratio between average region and phrase response magnitudes as
$
\rho_b
=
\frac{
\frac{1}{|\mathcal{R}_b|}
\sum_{r\in\mathcal{R}_b}\|s_r\|_2
}{
\frac{1}{|\mathcal{P}_b|}
\sum_{p\in\mathcal{P}_b}\|s_p\|_2
}.
$
We penalize scale mismatch outside margin $m$:
$
\mathcal{L}_{\mathrm{mag}}
=
\left[|\log\rho_b|-m\right]_+.
$
The margin permits moderate differences in overall activation magnitude between modalities.

\paragraph{Latent-level modality balance.}
Matching aggregate magnitude does not ensure that corresponding phrases and regions activate the same latent features. For matched pair $(p,\pi(p))$, let
$
u_j=[s_{\pi(p)}]_j,
\qquad
v_j=[s_p]_j
$
denote the visual and textual responses of latent $j$. We define its participation score as
$
\gamma_j
=
\operatorname{sg}\!\left[\max(u_j,v_j)\right],
$
where $\operatorname{sg}$ denotes stop-gradient, and retain the strongest participating latents
$
\mathcal{T}_p
=
\operatorname{TopK}_{K_{\mathrm{bal}}}(\gamma).
$
Their normalized participation weights are
$
\omega_j
=
\frac{\gamma_j}
{\sum_{j'\in\mathcal{T}_p}\gamma_{j'}},
$
and the visual share of latent $j$ is
$
\sigma_j
=
\frac{u_j}{u_j+v_j+\epsilon}.
$

We penalize only excessive modality imbalance:
$
\mathcal{L}_{\mathrm{bal}}
=
\frac{1}{|\mathcal{P}_b|}
\sum_{p\in\mathcal{P}_b}
\sum_{j\in\mathcal{T}_p}
\omega_j
\left[
\left|\sigma_j-\frac{1}{2}\right|-\mu
\right]_+^2.
$
The participation gate restricts this constraint to latents relevant to the matched content, while $\mu$ allows moderate modality specialization rather than forcing identical image and text responses.

\paragraph{Combined cross-modal objective.}
The complete phrase--region objective is
$
\mathcal{L}_{\mathrm{cross}}
=
\lambda_{\mathrm{al}}\mathcal{L}_{\mathrm{align}}
+
\lambda_{\mathrm{mag}}\mathcal{L}_{\mathrm{mag}}
+
\lambda_{\mathrm{bal}}\mathcal{L}_{\mathrm{bal}}.
$
Together, these terms align phrase and region representations while maintaining compatible response scale and shared latent participation across modalities.

\subsubsection{AuxK Activation for Dead-Latent Recovery}
\label{app:auxiliary}

We use the AuxK objective of \citet{gao2024scaling} to encourage inactive latents to re-enter the SAE dictionary. AuxK reconstructs the residual error using features that are inactive under the main TopK pathway, providing an auxiliary gradient to otherwise unused latents. In our setting, however, the choice of activation used in this auxiliary pathway is important because many inactive latents have negative pre-activations.

Using a standard ReLU causes such latents to receive zero auxiliary activation and therefore no recovery signal. A direct alternative is to use the raw signed pre-activations, but we found this unstable: negative responses can contribute with unrestricted magnitude through the decoder and substantially overshoot the reconstruction residual. In our experiments, the resulting auxiliary reconstruction magnitude was approximately $7$--$14\times$ larger than the residual it was intended to model, causing the auxiliary term to dominate the objective.

We also considered Softplus as a smooth alternative. However,
$\operatorname{softplus}(0)=\log 2$, so latents near the activation threshold contribute a nonzero constant even when their pre-activation is approximately zero. With a large dictionary, these small offsets accumulate through the decoder and introduce a systematic auxiliary reconstruction component. Softplus also alters the positive branch, making the auxiliary pathway inconsistent with the ReLU-based activation used by the main SAE.

We therefore use a leaky rectifier in the AuxK pathway,
\begin{equation}
\phi_{\mathrm{aux}}(x)
=
\begin{cases}
x, & x \ge 0,\\
\alpha x, & x < 0,
\end{cases}
\qquad
\alpha = 0.01.
\label{eq:aux-leaky-relu}
\end{equation}
This preserves the standard ReLU response for positive pre-activations while providing a small nonzero gradient to negative inactive features. The modification therefore leaves the positive auxiliary path unchanged but allows persistently negative latents to receive a recovery signal.

As a sanity check, we constructed a synthetic setting in which a subset of candidate auxiliary latents remained negative throughout the forward pass. Under ReLU, these latents produced zero auxiliary reconstruction and zero gradient. Replacing ReLU with the leaky rectifier yielded a finite auxiliary reconstruction and nonzero gradients for the same latents, confirming that the modification restores a recovery path without introducing the large responses observed with raw signed activations.

We use this leaky AuxK activation only in the auxiliary dead-latent recovery pathway; the main TopK encoding and reconstruction objective are unchanged.

\section{Implementation Details}
\label{app:implementation}

\paragraph{SAE configuration.}
Across both backbones, we use an SAE with $8192$ latents and TopK sparsity with $k=128$. Representations are extracted from a single decoder layer, with placement matched approximately by relative depth across models: layer $15$ of $28$ for Qwen3-VL-2B-Instruct and layer $19$ of $35$ for Gemma-4-E2B-it, corresponding to relative depths of $0.536$ and $0.543$, respectively. Inputs are limited to $512$ tokens, and images are resized to a maximum side length of $512$ pixels.

\paragraph{Gemma intervention depth.}
Gemma-4-E2B-it shares key--value representations across its upper decoder layers: layers above the key--value boundary reuse keys and values computed earlier in the network.
This architectural property does not affect experiments that intervene directly on all relevant token positions, but it prevents the patch-level causal grounding test from being evaluated at the primary $l=19$ SAE placement. 

\paragraph{Influence parameter scope.}
Gemma-4-E2B-it additionally contains an audio tower that is inactive for our image--text tasks. Parameters belonging exclusively to this unused modality receive no gradients and are therefore excluded from the influence-function parameter scope. All remaining active model parameters within the selected influence scope are included in the EK-FAC computation.

\subsection{Warm-Start Initialization}
\label{app:warmstart}

We train the structured SAE in two stages. Stage A first learns a stable sparse dictionary using reconstruction alone, after which Stage B initializes from these learned SAE weights and introduces the additional shaping objectives. The second stage therefore begins from a well-formed dictionary rather than learning reconstruction, sparsity, localization, and cross-modal alignment simultaneously.

This warm start is important for optimization stability. Training Stage B directly from scratch with strong phrase--region alignment ($\lambda_{\mathrm{al}}=0.1$) causes substantial latent collapse, with 3,371 latents ($41\%$) becoming inactive; adding load balancing does not prevent this failure. Increasing the amount of Stage-A pretraining progressively reduces collapse: 2,000 warm-up steps leave 1,927 dead latents ($24\%$), while 6,000 steps reduce this to 142 ($1.7\%$), close to the 56 dead latents obtained with the full Stage-A training. We therefore use the completed reconstruction-only Stage-A model to initialize Stage B throughout our experiments.

\subsection{Dataset Processing and Task Construction}
\label{sec:appendix_dataset_processing}

\paragraph{GQA.}
We construct GQA from the official balanced training and validation splits. To reduce redundancy from repeated questions about the same image, we retain at most one question per image using a fixed random seed. Training examples are sampled from \texttt{train\_balanced}, while validation and test examples are disjoint subsets of \texttt{val\_balanced}.

We use the original question verbatim as the prompt, without additional instructional scaffolding. This avoids introducing repeated boilerplate tokens that carry little example-specific information but can absorb attribution mass in latent and influence analyses. The target follows a \emph{short-then-full} format: the canonical short answer appears first, followed by GQA's provided full-answer sentence. The short answer preserves the standard task output, while the full sentence provides sufficient linguistic structure for token-level multimodal attribution.

\paragraph{CheXpert.}
We use frontal chest radiographs and consider five findings: Atelectasis, Cardiomegaly, Consolidation, Edema, and Pleural Effusion. For each finding, labels are mapped to three classes: positive labels indicate \emph{present}, uncertain labels remain \emph{uncertain}, and all remaining labels are treated as \emph{absent}. We sample examples to balance these three classes independently for each finding, using a fixed random seed. Validation patients are held out from training to prevent patient-level leakage.

Each example is prompted with the image and the question
\emph{``For this chest X-ray, what is the status of \{finding\}?''},
with three answer choices corresponding to absent, uncertain, and present. The prediction target begins with the corresponding answer letter, followed by the cleaned report impression. We use this \emph{impression-after} construction throughout the reported experiments, which preserves a concise classification target while providing clinically meaningful text for latent and attribution analysis.

\paragraph{VizWiz-Captions.}
We use the standard VizWiz caption annotations. We remove captions marked as rejected or precanned, as well as captions matching the dataset's precanned placeholder response. The prompt is fixed as
\emph{``Generate an accurate caption for this image.''}

During model training, the target is a valid human reference caption. For latent necessity and sufficiency analyses, we instead use the finetuned model's own generated caption as the target. This evaluates whether perturbing a latent recovers or disrupts the model's actual prediction, rather than measuring agreement with an alternative human reference.

\section{Additional Modality Analysis}
\label{app:modality_analysis}

Section~\ref{sec:exp:visual_evidence} studies how removing visual evidence changes
training-example and latent influence rankings. Here we examine whether the
strong image--text token-count imbalance itself biases the influence estimator.

\subsection{Modality Imbalance}

Vision--language sequences are strongly imbalanced in token count. Image tokens constitute approximately $78\%$ of a CheXpert sequence and $93\%$ of a VizWiz sequence. Because the influence computation aggregates over sequence positions, this raises a natural question: should image and text contributions be explicitly balanced? We study two possible corrections, acting on either the curvature estimate or the representation gradient.

\paragraph{Balancing the curvature estimate.}
A direct approach is to reweight token positions when estimating the EK-FAC factors so that image and text tokens contribute equal total mass. We interpolate between the standard and fully balanced estimators using
\[
w_{\alpha}
=
(1-\alpha)\mathbf{1}+\alpha w,
\qquad
\alpha\in[0,1],
\]
where $w$ assigns modality-balanced token weights. The weighting is applied as $\sqrt{w_{\alpha}}$ to the corresponding EK-FAC factors so that the resulting covariance receives weight $w_{\alpha}$.

This correction consistently degrades attribution quality. The CIDEr drop obtained by ablating the top $200$ attributed latents decreases from
$91.3\%$ at $\alpha=0$ to
$90.4\%$, $85.6\%$, $80.4\%$, and $77.8\%$ for
$\alpha=0.25,0.5,0.75,$ and $1$, respectively.
Thus, equalizing modality mass in the curvature does not reveal a better operating point; performance decreases monotonically as the correction is strengthened.

This behavior is not caused by the curvature correction being negligible. The eigenvalue-corrected spectrum is highly concentrated, with a median-layer participation ratio of $0.0003$ of the full dimension while $99.95\%$ of entries remain nonzero. Reweighting therefore substantially changes the inverse-curvature operator, but emphasizes directions that are less useful for the downstream attribution objective.

\paragraph{Balancing the representation gradient.}
An alternative is to balance modalities after curvature correction, directly in the representation-space influence vector. Let $s_{\mathrm{img}}$ and $s_{\mathrm{txt}}$ denote the contributions aggregated over image and text positions. We form
\begin{equation}
\tilde{s}_{\alpha}
=
\alpha
\frac{s_{\mathrm{img}}}{\|s_{\mathrm{img}}\|_2}
+
(1-\alpha)
\frac{s_{\mathrm{txt}}}{\|s_{\mathrm{txt}}\|_2},
\end{equation}
and restore the magnitude of the unrestricted influence vector,
\begin{equation}
s_{\alpha}
=
\tilde{s}_{\alpha}
\frac{\|s_{\mathrm{all}}\|_2}
     {\|\tilde{s}_{\alpha}\|_2}.
\label{eq:modbal}
\end{equation}
The final rescaling is important because influence depends on both alignment and magnitude: normalizing every example to unit norm would discard information about how strongly a training example affects the prediction.

In practice, this correction changes the ranking very little. On VizWiz with $\alpha=0.5$, the balanced and standard influence rankings have Spearman correlation $0.994$ over $60$ evaluation examples. The top-ranked training example is unchanged for every example, and the mean top-$8$ overlap is $1.00$.

\paragraph{Why modality balancing has little effect.}
The weak effect of gradient balancing follows from the structure of the learned latent space. Across latent dimensions, the image- and text-activation mass profiles have cosine similarity $0.73$, or $0.733$ when restricted to the $7{,}599$ of $8{,}192$ latents active in both modalities. Image and text therefore do not occupy approximately orthogonal subspaces; instead, they tend to activate many of the same latent directions in similar proportions.

Consistent with this observation, the cosine similarity between the modality-balanced vector and the original influence direction is
$0.944$, $0.988$, $0.999$, $0.973$, and $0.916$
for $\alpha=0,0.25,0.5,0.75,$ and $1$, respectively. In particular, the nominally balanced choice $\alpha=0.5$ is almost identical to the unrestricted direction.

This also helps explain why curvature-side balancing is ineffective. Although image tokens dominate numerically, the imbalance does not correspond to a distinct ``image direction'' that can simply be downweighted relative to an independent text direction. Equalizing token mass therefore perturbs the curvature without isolating a complementary textual subspace.

\paragraph{Interpretation.}
These results indicate that token-count imbalance should not automatically be interpreted as bias in the influence estimate. The sequence is highly imbalanced at the input level, but the learned latent basis substantially overlaps across modalities. Consequently, simple reweighting either leaves the attribution nearly unchanged or degrades its causal fidelity. We therefore use the unweighted estimator in our main experiments and report modality balancing as a negative ablation.

\subsection{Contrasting Influence With and Without Visual Evidence}
\label{app:visual_evidence}

We further analyze the image-removal experiment from
Sec.~\ref{sec:exp:visual_evidence}. For each test example, influence is computed
under matched conditions while changing only whether the
image is present. The text-only condition therefore intervenes on the
\emph{input}, rather than modifying the learned latent dictionary.
We compare
latent influence correlation, candidate-pool overlap, example-level influence
correlation over the union of active examples, and correlation restricted to
shared candidates. We additionally report top-8 attribution overlap as an
interpretable measure of ranking stability.

\begin{table}[t]
\centering
\caption{
Influence with and without visual evidence. Latent $\rho$ compares absolute
latent influence, example $\rho_{\mathrm{union}}$ compares training examples
active in either condition, and example $\rho_{\mathrm{shared}}$ is restricted
to candidates retrieved in both conditions.
}
\label{tab:visual_evidence_appendix}
\small
\setlength{\tabcolsep}{4pt}
\begin{tabular}{lccccc}
\toprule
Dataset / Model
& Pool J.
& Latent $\rho$
& Example $\rho_{\mathrm{union}}$
& Example $\rho_{\mathrm{shared}}$
& Top-8 \\
\midrule
GQA / Qwen
& $0.123$ & $0.599$ & $-0.564$ & $0.260$ & $0.073$ \\

GQA / Gemma
& $0.051$ & $0.543$ & $-0.755$ & 0.183 & $0.000$ \\

VizWiz / Qwen
& $0.021$ & $0.584$ & $-0.793$ & $0.279$ & $0.010$ \\

VizWiz / Gemma
& $0.052$ & $0.571$ & $-0.774$ & $-0.063$ & $0.000$ \\

CheXpert / Qwen
& $0.145$ & $0.737$ & $-0.606$ & $0.570$ & $0.031$ \\

CheXpert / Gemma
& $0.026$ & $0.795$ & $-0.797$ & $0.850$ & $0.000$ \\
\bottomrule
\end{tabular}
\end{table}

Across all six settings (Table~\ref{tab:visual_evidence_appendix}), removing the image preserves substantially
more agreement at the latent level ($\rho=0.543$--$0.795$) than at the
training-example level ($\rho=-0.797$--$-0.564$). Candidate pools also change
substantially, with Jaccard overlap of only $0.021$--$0.145$, while top-8
attribution overlap remains at or below $0.073$.

Correlation among candidates retrieved in both conditions is more variable,
ranging from $-0.095$ to $0.850$. This suggests that the most consistent effect
of image removal is a change in the \emph{support} of training-data
attribution---which examples are retrieved as influential---while subsequent
re-ranking of shared examples depends more strongly on the model and task.

\section{Additional Qualitative and Hierarchical Attribution Analysis}
\label{app:qualitative_attribution}

We provide additional examples complementing Sec.~\ref{sec:exp:representation} and
Sec.~\ref{sec:exp:qual}. We separately examine test-conditioned latent
decompositions and influential training examples to show how the attribution
hierarchy behaves across examples and tasks.
These examples complement the quantitative results in Table~\ref{tab:latent_quality_generalization} by showing how the improved latent structure appears at the level of individual features. Across tasks, our objective produces latents that are more clearly associated with useful semantic terminology and whose visual support is more concentrated on the corresponding image evidence.

\begin{figure}[t]
\centering
\includegraphics[width=0.6\linewidth]{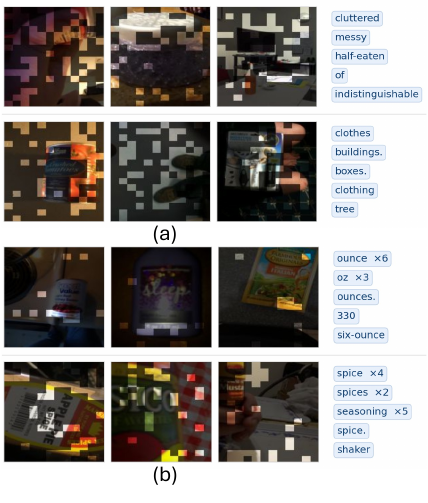}
\caption{\textbf{Example latent comparison on VizWiz.}
(a) Base SAE. (b) Our approach. Our latent exhibits a clearer semantic association and more spatially coherent visual support, while the base latent activates more diffusely across the image.}
\label{fig:vizwiz_latents}
\end{figure}

Figure~\ref{fig:vizwiz_latents} provides a corresponding comparison on VizWiz. The base SAE exhibits broader and less coherent visual support, whereas our objective produces latents with more interpretable semantic associations and activation concentrated on the image regions relevant to those semantics.

\begin{figure}[t]
\centering
\includegraphics[width=0.6\linewidth]{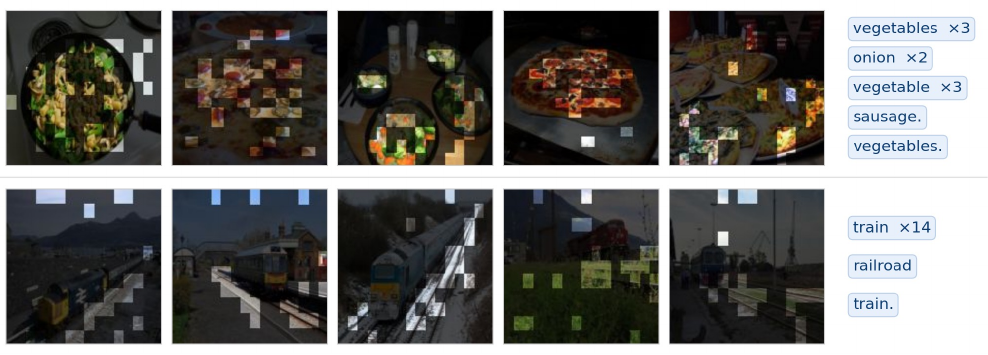}
\caption{\textbf{Example latents on GQA.}
Representative latents learned by our approach exhibits a meaningful semantic association together with localized activation over the corresponding visual evidence.}
\label{fig:gqa_latents}
\end{figure}

Finally, Fig.~\ref{fig:gqa_latents} shows representative GQA latents. Despite the different task structure, the learned features remain semantically interpretable and spatially concentrated on the image evidence associated with its textual meaning. Together, these examples illustrate that the improved localization and semantic quality observed quantitatively are also apparent across individual latents, datasets, and task domains.

\subsection{Detailed Analysis of Influential Test Latents}
\label{app:vizwiz_test_latents}

We further examine the latent decomposition of two individual VizWiz test examples to understand what information is represented beyond the aggregate latent-quality metrics. Rather than considering only the highest-scoring latent, we inspect a broader set of latents activated by each test image and compare their associated textual semantics with the image regions on which they respond. These examples show that the learned representation captures not only the primary object relevant to the prediction, but also complementary object parts, interactions, and scene context.

\begin{figure}[t]
\centering
\includegraphics[width=0.7\linewidth]{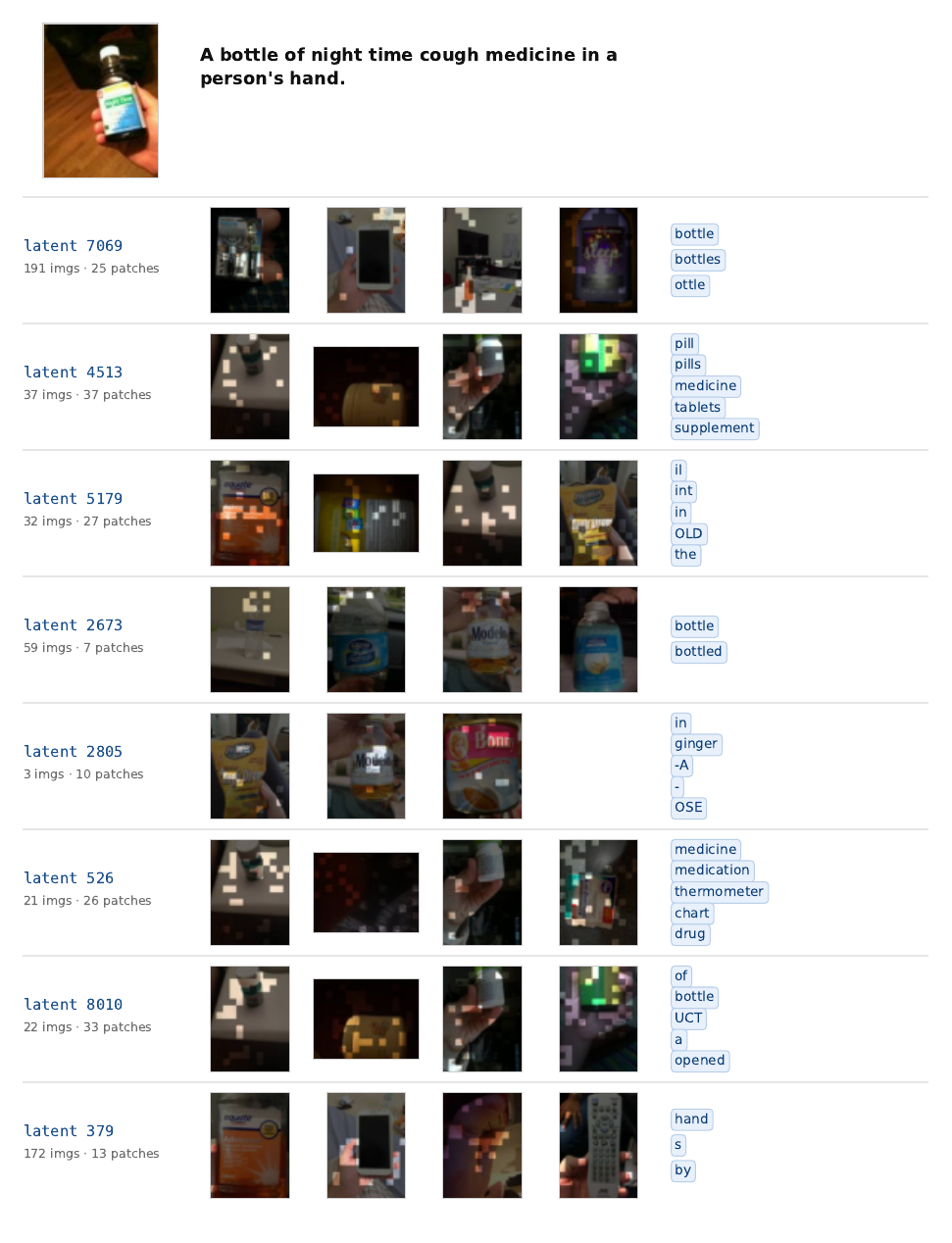}
\caption{\textbf{Detailed latent decomposition for a VizWiz medicine example.}
Representative test latents capture complementary evidence including the \emph{bottle}, \emph{medicine}-related semantics, and the \emph{hand holding the medicine}. The localized activations show that the representation separates object identity, semantic function, and relevant object--person interaction while remaining grounded in the corresponding image regions.}
\label{fig:vizwiz_test_medicine}
\end{figure}

Figure~\ref{fig:vizwiz_test_medicine} shows a test example centered on a medicine bottle. The highest-response latents recover several mutually consistent aspects of the scene, including concepts related to \emph{bottle} and \emph{medicine}. Importantly, other latents localize to the \emph{hand holding the medicine}, capturing the interaction between the object and the person rather than simply firing on the bottle in isolation. This decomposition is particularly relevant to the test prediction: the model represents the object identity, its functional category, and the local visual evidence establishing how the object appears in the scene. The resulting latent set therefore provides a substantially richer account of the example than a single global visual feature.

\begin{figure}[t]
\centering
\includegraphics[width=0.7\linewidth]{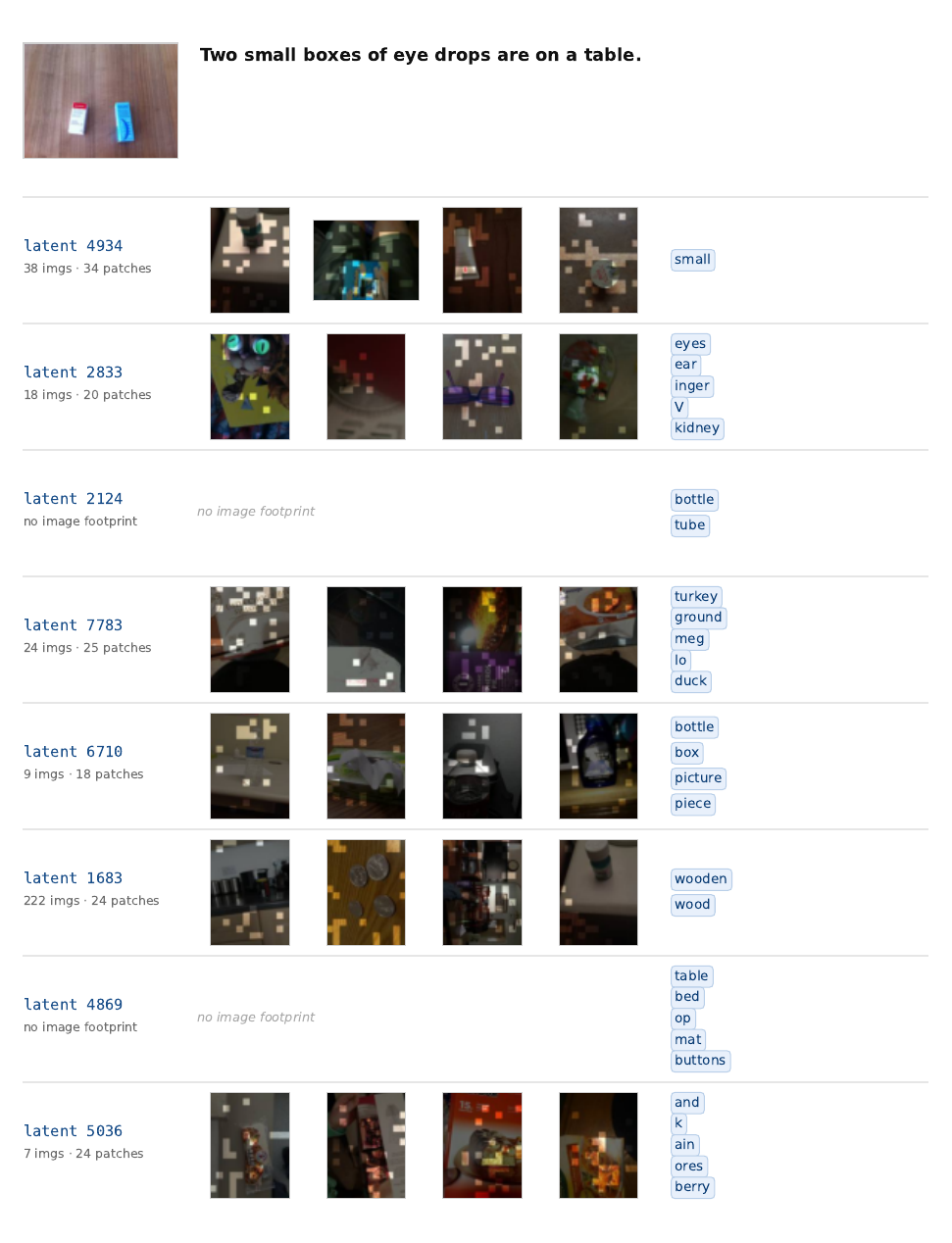}
\caption{\textbf{Detailed latent decomposition for a VizWiz eye-drop example.}
The learned latents recover the central semantics of the test image, including \emph{eyes}, \emph{eye drops}, and \emph{bottle}, while also representing localized contextual evidence such as the \emph{wooden table} on which the object appears. This example illustrates how distinct latents jointly encode both answer-relevant concepts and faithful scene context.}
\label{fig:vizwiz_test_eyedrops}
\end{figure}

A second example, shown in Fig.~\ref{fig:vizwiz_test_eyedrops}, contains a bottle of eye drops placed on a wooden table. Here, the latent decomposition again separates multiple visually and semantically meaningful components. Several latents correspond directly to the principal concepts needed to interpret the image, including \emph{eyes}, \emph{eye drops}, and \emph{bottle}. At the same time, other latents capture contextual properties of the scene, such as the \emph{wood} or wooden surface beneath the object. This contextual latent is notable because the test image itself contains a prominent wooden tabletop: although this evidence is not the primary answer concept, it faithfully reflects a real and localized component of the input rather than an unrelated semantic association.

Together, these examples highlight two useful properties of the learned latent space. First, semantically related concepts are decomposed across distinct but complementary latents rather than being collapsed into a single diffuse feature. Second, the associated visual support remains faithful to the specific test image: object latents localize to the relevant objects or parts, interaction latents capture relationships such as the hand holding the medicine, and contextual latents recover visible scene attributes such as the wooden tabletop. This fine-grained decomposition makes the latent representation useful not only for identifying \emph{what} semantic evidence is present, but also for tracing \emph{where} that evidence appears and how multiple pieces of evidence jointly support the model's prediction.

\subsection{Qualitative Analysis of Influential Training Examples}
\label{app:train_example_analysis}

We next inspect the training examples assigned the highest influence scores for individual test predictions. These examples provide a complementary view to the latent-level analysis: rather than asking which concepts within an example matter, we examine whether the retrieved training examples themselves contain evidence that is semantically consistent with the test query and prediction. Across VizWiz, GQA, and CheXpert, the most influential examples exhibit
prediction-relevant correspondence with the test instance, ranging from
direct visual or answer-level agreement to related clinical evidence across
diagnostic labels.

We additionally expand the VizWiz medicine example introduced in the main text (Fig.~\ref{fig:influence_chain}) to inspect the influential training examples in greater detail. The highest-ranked examples capture several complementary aspects of the test instance rather than simply repeating the same object. Some provide direct semantic support through examples involving \emph{medicine} or \emph{cough medicine}, while others contain visually related containers and packaging that help distinguish the target object from alternative possibilities. Additional examples reflect contextual properties present in the query image, including the wooden tabletop on which the medicine is shown. Together, these examples illustrate that the influence ranking captures both answer-relevant semantic evidence and visually grounded contextual evidence, providing a more nuanced explanation of the prediction than simple nearest-neighbor similarity.

\begin{figure}[t]
\centering
\includegraphics[width=0.7\linewidth]{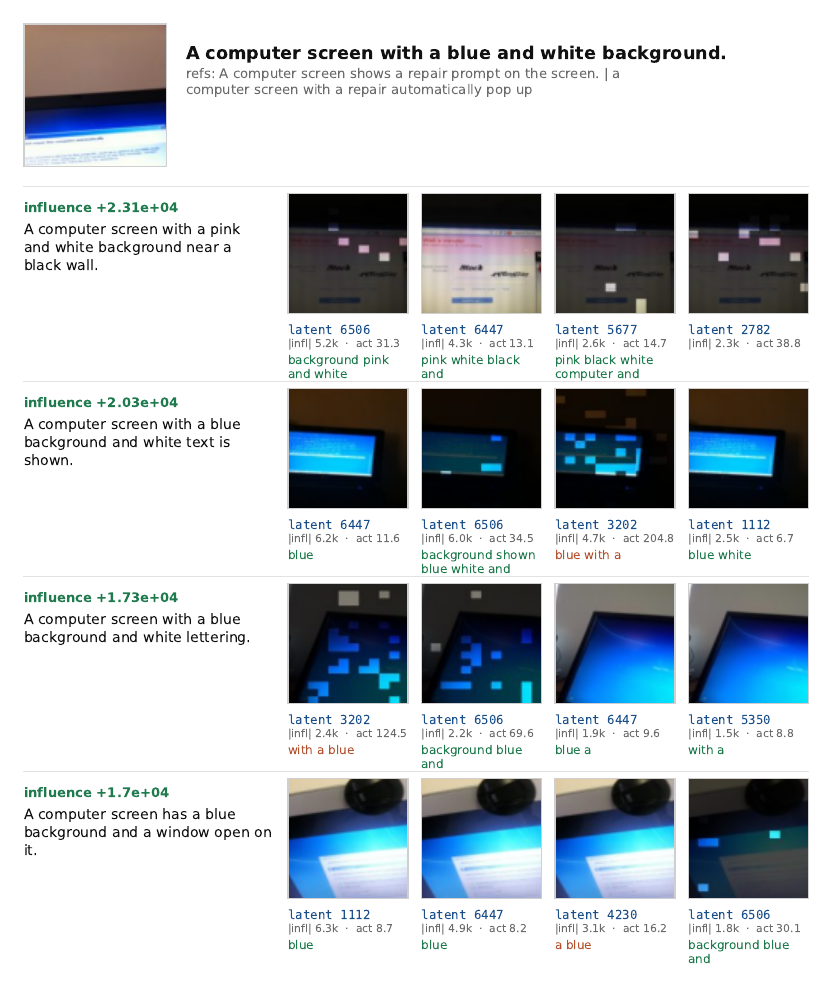}
\caption{\textbf{Influential training examples for a VizWiz screen query.}
The test example contains a screen, and the highest-influence training examples consistently exhibit closely related screen or display content. This qualitative consistency shows that influential examples reflect the salient visual semantics of the test input.}
\label{fig:vizwiz_train_screen}
\end{figure}

Figure~\ref{fig:vizwiz_train_screen} shows a VizWiz test example centered on a screen. The highest-ranked training examples consistently contain closely related screen or display content, indicating that the influence estimator retrieves examples with strong visual and semantic correspondence to the test input. Despite variation in the surrounding scene, the shared screen-related content remains prominent across the influential examples.

\begin{figure}[t]
\centering
\includegraphics[width=0.7\linewidth]{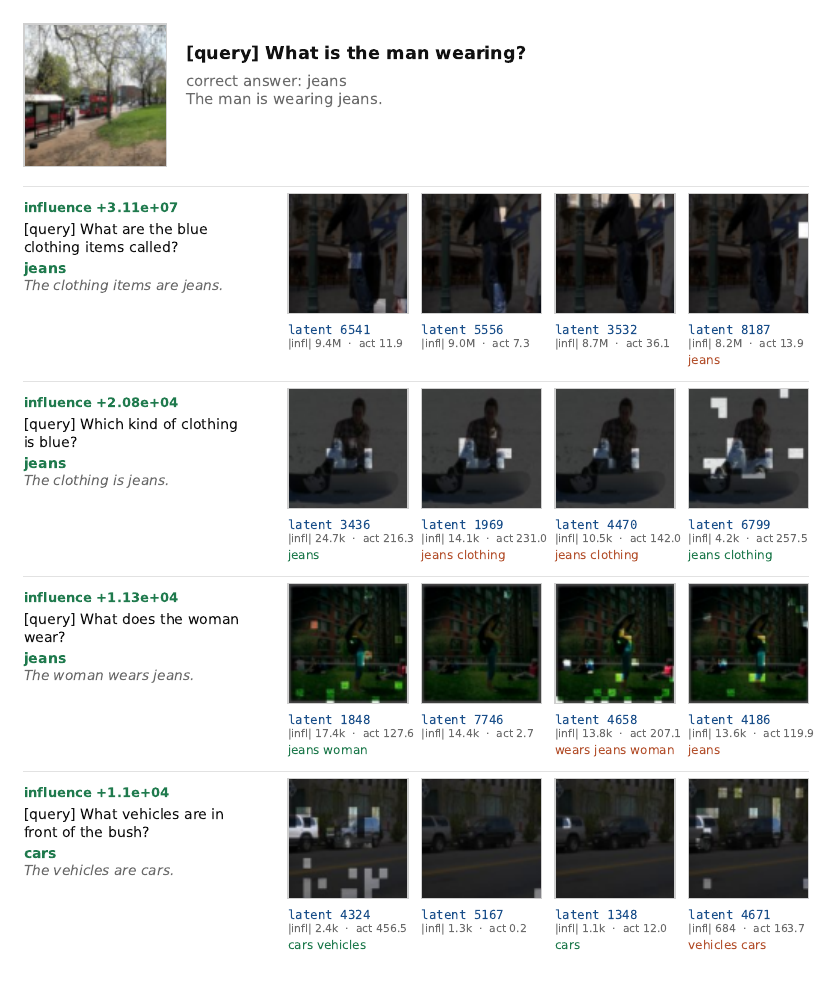}
\caption{\textbf{Influential training examples for a GQA clothing query.}
The test question asks what the man is wearing, with \emph{jeans} as the predicted answer. The top three influential training examples are all associated with \emph{jeans}, directly matching the answer semantics of the test query.}
\label{fig:gqa_train_jeans}
\end{figure}

Figure~\ref{fig:gqa_train_jeans} provides a GQA example in which the query asks about what a man is wearing. The model's answer is \emph{jeans}, and the three most influential training examples are likewise associated with \emph{jeans}. The agreement is therefore not only visual but directly aligned with the answer concept, suggesting that the prediction is supported by training examples encoding the same attribute queried at test time.

\begin{figure}[t]
\centering
\includegraphics[width=0.7\linewidth]{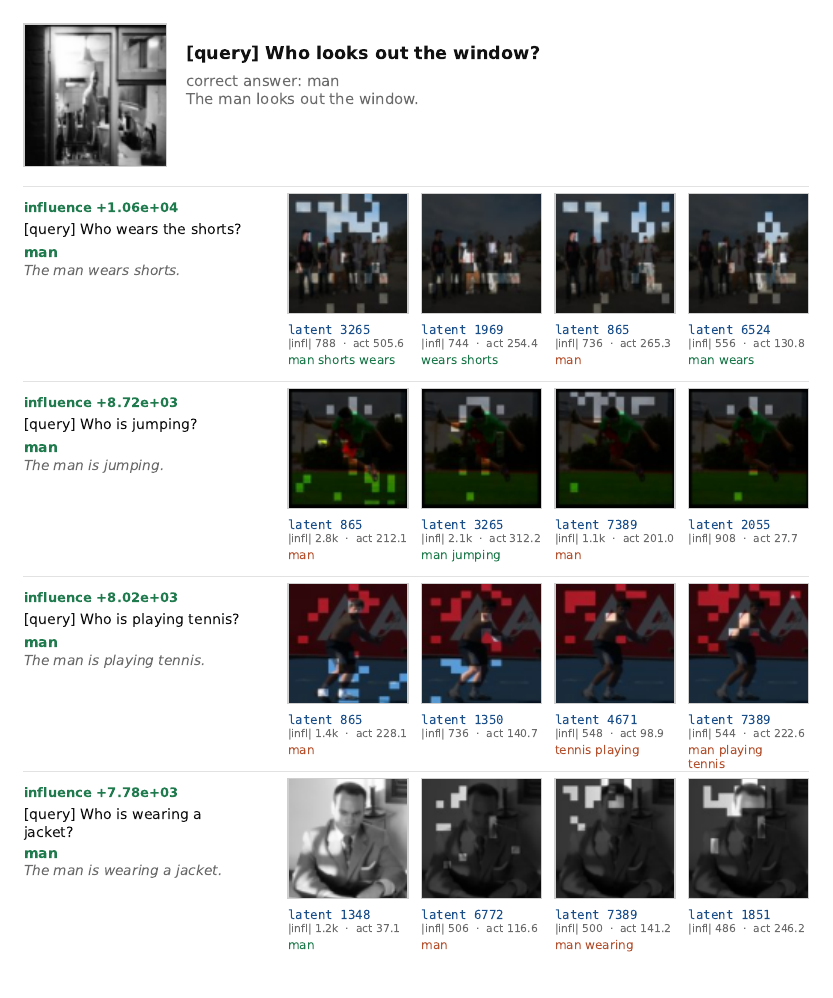}
\caption{\textbf{Influential training examples for a GQA identity query.}
The test question asks who is looking out the window, with \emph{man} as the answer. The highlighted answers in the highest-influence training examples are likewise \emph{man}, demonstrating strong agreement between the training supervision and the test prediction.}
\label{fig:gqa_train_man}
\end{figure}

A second GQA example is shown in Fig.~\ref{fig:gqa_train_man}. Here, the test question asks who is looking out the window, for which the answer is \emph{man}. The highlighted answers of the most influential training examples are also \emph{man}. This agreement indicates that the influence ranking identifies training examples whose supervision is directly consistent with the semantic decision made for the test instance.

We further examine influential training examples on CheXpert, where the
prediction task differs from the open-ended settings above. Here, the
highest-influence examples need not share the same queried finding as the
test example. Instead, they often contain related radiographic patterns or
report semantics, providing a complementary view of how training examples
support clinical predictions.

\begin{figure}[t]
\centering
\includegraphics[width=0.7\linewidth]{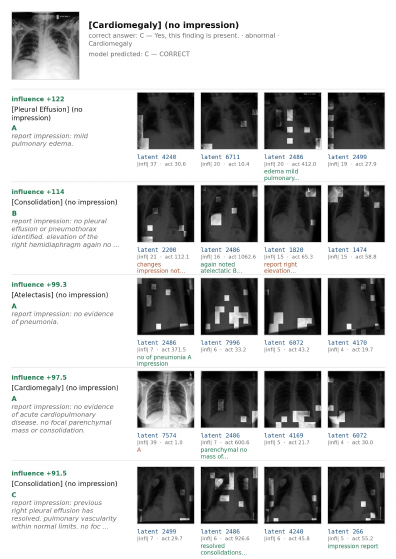}
\caption{\textbf{Influential training examples for a CheXpert cardiomegaly
prediction.}
The correctly predicted cardiomegaly case receives support from training
examples spanning multiple queried findings, including cardiomegaly and
other cardiopulmonary abnormalities. This illustrates that influence is not
determined solely by exact diagnostic-label matching.}
\label{fig:chexpert_train_cardio}
\end{figure}

Figure~\ref{fig:chexpert_train_cardio} shows a correctly predicted
\emph{cardiomegaly} example. The influential training set spans several
queried findings, including pleural effusion, consolidation, atelectasis,
and cardiomegaly itself. Thus, the ranking is not explained simply by
retrieving examples with the same diagnostic label. Rather, the influential
examples capture overlapping cardiopulmonary evidence, while the localized
latent activations identify the image regions through which this evidence
contributes to the prediction.

\begin{figure}[t]
\centering
\includegraphics[width=0.7\linewidth]{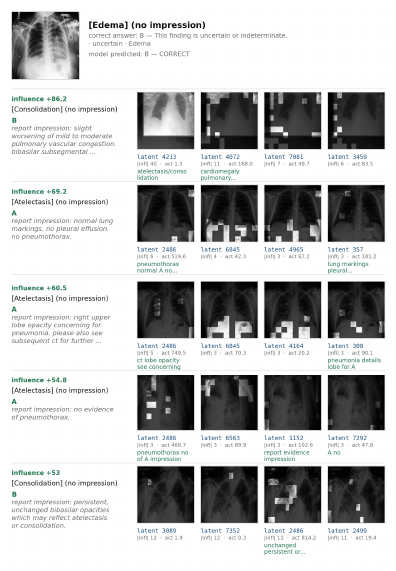}
\caption{\textbf{Influential training examples for a CheXpert edema
prediction.}
The highest-influence example is queried for consolidation but its report
describes pulmonary vascular congestion, providing evidence closely related
to edema. The ranking therefore captures clinically relevant cross-label
correspondence rather than only exact label agreement.}
\label{fig:chexpert_train_edema}
\end{figure}

The edema example in Fig.~\ref{fig:chexpert_train_edema} makes this
cross-label behavior particularly clear. Although the most influential
training example is queried for \emph{consolidation}, its report describes
mild-to-moderate pulmonary vascular congestion, evidence that is
semantically closely related to the edema prediction. Other highly
influential examples similarly contain pulmonary opacities and related
abnormal chest findings. The retrieved examples therefore reflect the
underlying evidence relevant to the prediction rather than merely the
surface identity of the queried label.

\begin{figure}[t]
\centering
\includegraphics[width=0.7\linewidth]{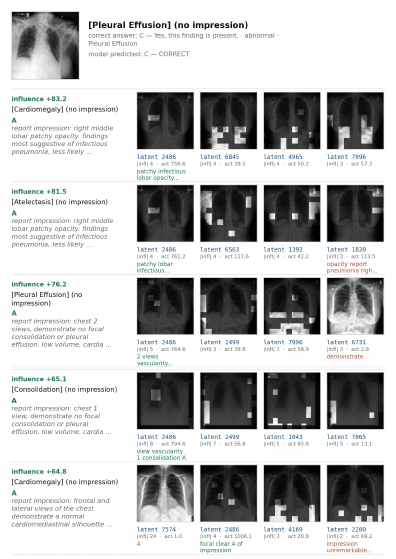}
\caption{\textbf{Influential training examples for a CheXpert pleural
effusion prediction.}
The influential set contains both a directly matching pleural-effusion
example and examples associated with other thoracic findings, demonstrating
a mixture of direct and cross-label support for the prediction.}
\label{fig:chexpert_train_effusion}
\end{figure}

A similar pattern appears for the \emph{pleural effusion} prediction in
Fig.~\ref{fig:chexpert_train_effusion}. A directly matching pleural-effusion
example appears among the highest-ranked training examples, while other
influential cases are associated with cardiomegaly, atelectasis, and
consolidation. The combination of direct and cross-label support suggests
that the influence estimator captures shared radiographic evidence across
related findings instead of reducing attribution to examples carrying the
same task label.

Together, these cases show that high-influence training examples are not
arbitrary nearest examples: they preserve prediction-relevant information
from the test instance. The VizWiz example exhibits strong visual-semantic
correspondence around the screen concept, while the GQA cases show direct
agreement at the answer level for \emph{jeans} and \emph{man}. In CheXpert,
the correspondence is more structured: influential examples may cross
diagnostic labels while retaining related radiographic or report evidence.
This provides qualitative evidence that the influence ranking surfaces
training examples that are meaningfully aligned with the model's test-time
decision rather than simply matching superficial labels or visual
similarity.

\section{Comparison with Pooled Multimodal SAE Baselines}
\label{app:pooled_sae_baselines}

We compare \method{} against two recent multimodal SAE approaches that learn
aligned image--text representations: JSAE~\citep{shu2026joint} and
SPARC~\citep{nasirisarvi2026sparc}. These methods provide natural controls for
whether a generic shared or aligned multimodal latent space is sufficient for
our attribution setting.

\paragraph{Experimental setup.}
All methods are trained on the same VizWiz activations from
Qwen3-VL-2B-Instruct at layer 15. We use the same underlying VLM representations
so that the comparison isolates the SAE objective. For SPARC, we use its
published Global-TopK and cross-reconstruction formulation. For JSAE, we
implement its separate image/text encoders and decoders with cosine alignment
between the pooled modality codes. We use per-modality normalization and
$\lambda_s=1.0$, which places JSAE near its reported sparsity regime rather than
evaluating it at the substantially denser operating point obtained with the
original hyperparameter on our activations.

\begin{table}[t]
\centering
\caption{
Comparison with pooled multimodal SAE baselines on VizWiz with Qwen.
Balance is computed over selective live latents. Footprint measures the
fraction of the image patch grid activated by a latent, and contiguity is
normalized against a size-matched random support. Lower footprint,
contiguity, and word concentration indicate more selective features.
}
\label{tab:pooled_sae_baselines}
\small
\setlength{\tabcolsep}{4pt}
\begin{tabular}{lrrrrrr}
\toprule
Method
& Alive
& Balance $\uparrow$
& Footprint $\downarrow$
& Contig. $\downarrow$
& Word conc. $\downarrow$
& Img. tok./fire $\downarrow$ \\
\midrule
\method{}
& 8156
& \textbf{26.7}
& \textbf{8.3\%}
& \textbf{0.548}
& \textbf{0.219}
& \textbf{3.30} \\
JSAE
& 99
& 3.3
& 100\%
& 1.000
& 0.552
& 192 \\
SPARC
& 69
& 0.0
& 100\%
& 1.001
& 0.992
& 192 \\
\bottomrule
\end{tabular}
\end{table}

\paragraph{Global alignment does not imply localized shared features.}
Table~\ref{tab:pooled_sae_baselines} shows a substantial difference in the
granularity of the learned representations. JSAE and SPARC can align
image- and text-level representations, but their active visual features span
essentially the full image. In contrast, \method{} retains thousands of live
features while producing substantially more selective and spatially localized
support. The distinction is particularly important for our use case: an
example-level shared concept is insufficient to determine which image patch or
text phrase carries a training example's influence.

This follows directly from the alignment objectives. JSAE aligns one
mean-pooled image code with one mean-pooled text code for each example, while
SPARC likewise learns correspondence at the representation-stream level.
Neither objective defines a phrase--patch correspondence. In \method{}, the
representation remains position-resolved and phrase--region supervision
explicitly determines which localized visual and linguistic evidence should
share latent structure.

\paragraph{Why pooled dictionaries cannot simply be applied per token.}
One possible alternative is to take a pooled multimodal dictionary and insert it
at every image and text position, thereby creating the position-resolved
representation required by latent influence. We test this directly by replacing
the residual-stream representation with each dictionary's per-token
reconstruction and measuring the resulting language-model loss.

\begin{table}[t]
\centering
\caption{
Effect of inserting each dictionary position-wise into the VLM forward path.
Mean cross-entropy is evaluated over 60 VizWiz validation examples.
$\Delta$ is relative to \method{}.
}
\label{tab:pooled_sae_splice}
\small
\begin{tabular}{lcc}
\toprule
Forward path & Mean CE $\downarrow$ & $\Delta$ \\
\midrule
Identity (no SAE) & 1.848 & $-0.180$ \\
\method{} & 2.027 & 0 \\
JSAE & 9.464 & $+7.44$ \\
Mean-broadcast control & 10.462 & $+8.43$ \\
SPARC & 12.691 & $+10.66$ \\
\bottomrule
\end{tabular}
\end{table}

As shown in Table~\ref{tab:pooled_sae_splice}, the pooled dictionaries do not
preserve the VLM computation when applied position-wise at our intermediate
layer. JSAE approaches the mean-broadcast control, while SPARC performs worse.
This does not imply that these methods fail at their intended pooled
representation or steering objectives; rather, their learned unit of analysis
is not interchangeable with a token- or patch-level SAE inserted into the
forward path.

This distinction matters for latent-mediated influence. Our attribution is
computed through the SAE-mediated forward pass, so the prediction must depend
on the same position-resolved latent variables that are subsequently attributed.
A pooled representation can instead be analyzed in parallel, but then the
model loss does not depend on those codes and their gradients cannot mediate
the prediction. \method{} is therefore designed specifically to preserve the
forward computation while providing a shared latent basis at the phrase,
token, patch, and region levels needed for grounded influence decomposition.

\paragraph{Takeaway.}
These comparisons isolate the difference between \emph{cross-modal alignment}
and an \emph{attribution-compatible multimodal basis}. Existing pooled
multimodal SAEs can learn shared image--text concepts, but do not natively
provide the position-resolved, spatially localized representation required to
trace training-example influence to specific linguistic and visual evidence.

\end{document}